\documentclass[letterpaper, 10pt, conference]{ieeeconf}

\let\labelindent\relax

\usepackage{bm}
\usepackage{svg}
\usepackage{pgf}
\usepackage{soul}
\usepackage{cite}
\usepackage{tikz}
\usepackage{bbold}
\usepackage{color}
\usepackage{amsmath}
\usepackage{amssymb}
\usepackage{wrapfig}
\usepackage{booktabs}
\usepackage{pdfpages}
\usepackage{graphicx}
\usepackage{titlesec}
\usepackage{amsfonts}
\usepackage{multirow}
\usepackage{hyperref}
\usepackage{subfiles}
\usepackage{subfigure}
\usepackage{algorithm}
\usepackage{dblfloatfix}
\usepackage[table]{xcolor}
\usepackage[inline]{enumitem}
\usepackage[noend]{algpseudocode}
\usepackage[mode=buildnew]{standalone}
\usepackage[font=small, belowskip=-0.5\baselineskip]{caption}

\hypersetup{
  colorlinks=true,
  linkcolor=black,     
  urlcolor=blue
}

\graphicspath{{./figures/}}

\usetikzlibrary{arrows,arrows.meta,patterns,decorations.markings,backgrounds,fit,positioning,shapes.callouts}

\newtheorem{theorem}{Theorem}[section]

\newtheorem{corollary}{Corollary}[theorem]
\newtheorem{defn}{Definition}
\newtheorem{remark}{Remark}

\titlespacing\section{0pt}{4pt}{4pt}
\titlespacing\subsection{0pt}{2pt}{2pt}
\titlespacing\subsubsection{0pt}{2pt}{2pt}

\newcommand{\ignore}[1]{}
\newcommand{\red}[1]{\textcolor{red}{#1}}

\newcommand{\mypara}[1]{\medskip\noindent\textbf{#1}:\hspace*{1ex}}

\definecolor{unassisted}{HTML}{A1C9F4}
\definecolor{assisted}{HTML}{FFB482}

\IEEEoverridecommandlockouts

\title{\LARGE \bf
Screw Geometry Meets Bandits: Incremental Acquisition of Demonstrations to Generate Manipulation Plans
}

\author{Dibyendu Das$^{1}$, Aditya Patankar$^{2}$, Nilanjan Chakraborty$^{2}$, C. R. Ramakrishnan$^{1}$, and I. V. Ramakrishnan$^{1}$
\thanks{$^{1}$Dept. of Computer Science, 
        Stony Brook University, USA.
        {\tt\small \{didas, cram, ram\}@cs.stonybrook.edu.}}%
\thanks{$^{2}$Dept. of Mech. Engg., 
        Stony Brook University, USA.
        {\tt\small \{aditya.patankar, nilanjan.chakraborty\} @stonybrook.edu.}}%
}

\begin{document}

\maketitle

\pagestyle{empty}

\begin{abstract}
In this paper, we study the problem of methodically obtaining a sufficient set of kinesthetic demonstrations, one at a time, such that a robot can be confident of its ability to perform a complex manipulation task in a given region of its workspace.  
Although programming by demonstration has been an active area of research,  
the problems of checking whether a set of demonstrations is sufficient and systematically seeking additional demonstrations have remained open. 
We present an approach for the robot to incrementally and actively ask for new demonstration examples, one at a time, until the robot can assess with high confidence that it can perform the task successfully. Our approach uses
\begin{enumerate*}[label=(\roman*)]
  \item a screw geometric representation of motion to generate manipulation plans from demonstrations, which makes the sufficiency of a set of demonstrations \emph{measurable};
  \item a sampling strategy based on PAC-learning from multi-armed bandit optimization to evaluate the robot's ability to generate manipulation plans in a subregion of its task space; and
  \item a heuristic to seek additional demonstration from areas of weakness.
\end{enumerate*}
We present results of a user study conducted with $22$ participants (without any background in robotics) on two example manipulation tasks, namely pouring and scooping, to assess the utility and usability of our approach. The results show that a handful of examples (fewer than $10$) were needed to successfully teach the robot to plan tasks. A short video supplement is available on YouTube: \href{https://youtu.be/KbAPgIouIvo}{https://youtu.be/KbAPgIouIvo}
\end{abstract}


\section{Introduction}
\label{sec:intro}

The ability of robots to go beyond pick-and-place tasks and perform complex manipulation tasks is key to their use in many application areas, including flexible manufacturing, service robotics, and assistive robotics. In complex manipulation tasks, such as opening drawers, doors, pouring, scooping, etc., it is crucial for the motion of the end effector to follow geometric path constraints during task execution. It is often difficult to manually specify the complete set of constraints for each task. 

One popular approach is to generate manipulation plans (a sequence of configurations of robot manipulators, such as its arms and grippers) directly based on a set of human-provided \emph{demonstrations}.
Such demonstrations can be obtained from video, teleoperation,  kinesthetic manipulation of the robot's end effector, or simulations in virtual or augmented reality, e.g., see~\cite{argall2009survey, chernova2014robot, calinon2010learning, Fischer2016ACO, perez2017c, herrero2021understanding, bahl2021hierarchical, yu2022user, fu2024mobile, zitkovich2023rt, peternel2022after}.

However, the following challenges hamper the use of current demonstration-based methods for practical applications:
\begin{enumerate*}[label=(\alph*)]
    \item the high cost of obtaining human-provided demonstrations, especially for kinesthetic or teleoperated demonstrations, which implies fewer demonstrations;
    \item the opaqueness of the learning process -- there is no way for the teacher to know which (if any) additional demonstrations would be useful.
\end{enumerate*}
Thus, despite this rich history,  the problem of evaluating the \emph{sufficiency} of a given set of demonstrations and systematically \emph{seeking} additional human-provided demonstrations has remained open.

\begin{figure}
    \centering
    \includegraphics[width=0.7\linewidth]{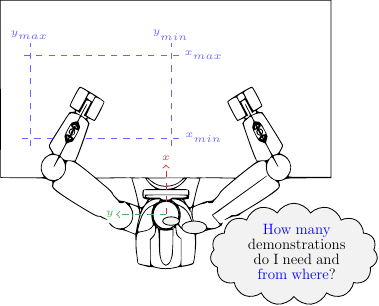}
    \caption{Schematic sketch of a robot working in a table-top environment. The work area is indicated by the dashed rectangle.}
    \label{Fig:problem}
\end{figure}

\subsection{Our Setting}
We focus on manipulation tasks with rigid objects in a tabletop environment where all task-relevant objects are located within a \emph{work area} (see Fig.~\ref{Fig:problem}).  The work area lies entirely within the robot's reachable workspace.
We assume that \emph{positions} and \emph{orientations} (i.e. \emph{poses}) of task-relevant objects are known, possibly informed by vision and other sensor systems. Demonstrations are provided \emph{kinesthetically}, where a human teacher holds the robot's end effector to perform the demonstrated task. Although resource-intensive, kinesthetic teaching is generally considered more usable and has a higher task success rate~\cite{argall2009survey}. We restrict our attention to path-based task constraints (e.g. pulling a door by rotating it around its axis, holding a cup upright until it is rotated to pour out its contents into a bowl).

\subsection{Our Approach}
\mypara{Preserving Task Constraints}
Clearly, without an explicit specification of task constraints, it is impossible to determine whether a given manipulation plan satisfies the constraints. This encapsulates the fundamental difficulty in attaching any rigorous measure of \emph{sufficiency} to a set of demonstrations.  The key aspect of our approach that helps us circumvent this difficulty is that we generate manipulation plans from a kinesthetic demonstration via screw-linear interpolation, ScLERP~\cite{laha2021point, mahalingam2023human}.  
The manipulation plans generated by ScLERP combined with the Jacobian-pseudoinverse~\cite{sarker2020screw} implicitly preserve the task-related constraints present in the demonstration (see \S\ref{sec:related}). 
Since the satisfaction of task constraints is ensured, a demonstration is sufficient for a given task instance if a plan satisfying the joint limits of the manipulator can be generated.
This gives us a concrete way to measure the sufficiency of a set of demonstrations.

\mypara{Estimating Sufficiency}
Let $n$ denote the number of task-relevant objects. The pose of each object is an element of $SE(3)$, the group of rigid body configurations or motions.
Let $\mathcal{X} \subseteq SE(3)^n$ be a set of possible poses of the $n$ task-relevant objects. Each element of $\mathcal{X}$ is a \emph{task instance}. Given $\mathcal{X}$, and a threshold $\beta \in [0,1]$, we say that a set of demonstrations is sufficient if the probability that we can generate successful manipulation plans for task instances uniformly drawn from $\mathcal{X}$ exceeds $\beta$. Note that such a probability-based measure is, by itself, \emph{coarse}: Although a set of demonstrations may be sufficient for $\mathcal{X}$, there may be regions within $\mathcal{X}$ where we cannot generate any successful plan. To identify such regions of weakness, we develop a procedure based on PAC (``Probably Approximately Correct'') learning techniques from the ``Multi-armed Bandit optimization'' literature (see \S\ref{sec:bandit-formulation}). \ignore{
We partition $\mathcal{X}$ into a finite set of disjoint regions $\mathcal{X}_j: 1 \leq j \leq K$ for a given natural number $K$.  We draw samples from $\mathcal{X}_j$ to estimate the probability of generating successful manipulation plans for task instances in the $j$-th region.} PAC-learning in this setting lets us determine the minimum number of samples required in each region to obtain probability estimates with high confidence~\cite{even2002pac}.

\mypara{Incremental Demonstration Acquisition}
A region with a low estimated probability of generating successful plans is a candidate for the next demonstration. With this idea at its core, we have built a system with a user interface that incrementally seeks demonstrations until $\mathcal{X}$ is covered, i.e., we can successfully generate manipulation plans with high probability within each subregion. The salient aspect of this system is that it permits non-experts to provide demonstrations. The effectiveness of the system in reducing the number of demonstrations was evaluated with a systematic user study with $22$ non-expert participants providing demonstrations for two complex manipulation tasks, namely \textbf{scooping} and \textbf{pouring}. Thus, \emph{we present an approach to build end-to-end systems with rigorous performance assurances that allows non-roboticists to program robots through intuitive kinesthetic demonstrations}.

\subsection{Key Contributions of this Work}
\begin{enumerate}
\item We define a concrete measure of \emph{sufficiency} of a set of demonstrations from which we can generate manipulation plans for a given set of task instances (\S\ref{sec:solution-overview}).
\item We provide a sampling procedure to estimate the sufficiency measure for subregions of the given set of task instances (\S\ref{sec:solution}).
This provides a basis for seeking additional demonstrations from the human teacher.
\item We present a systematic user study with \emph{non-experts} to assess the utility of our approach in reducing the demonstration burden (\S\ref{sec:user-study}).
\end{enumerate}

\noindent
The results of the user study show that, using our approach, fewer than 10 demonstrations are sufficient to generate successful plans over a specified work area compared to arbitrary acquisition of demonstrations (\S\ref{sec:user-study}). We also performed simulation experiments to estimate distributions of the number of sufficient demonstrations for the two tasks (\S\ref{sec:simulation-results}).

We begin the technical development in this article with a more detailed description of the closely related work.

\section{Related Work}
\label{sec:related}

Manipulation tasks, such as pouring from a container or opening a hinged door, involve path constraints on motion. Such constraints are hard and often impossible to specify generically with soundness or completeness guarantees. Learning from Demonstrations (LfD)~\cite{chernova2014robot, billard2016learning, calinon2007onlearning} avoids this problem by using a set of demonstrations provided \emph{a priori} to generate a constrained motion. This naturally raises two questions:
\begin{enumerate*}
  \item whether motion plans generated from demonstrations indeed satisfy the implicit constraints of the given task, and 
  \item whether a set of demonstrations are sufficient to generate robust motion plans for tasks in a specified work area.
\end{enumerate*}

\subsection{Motion planning from demonstrations}

There are a wide variety of techniques for generating motion plans from demonstrations,  including probabilistic models based on hidden Markov models (HMM)~\cite{kulic2012incremental}, Gaussian mixture models (GMM)~\cite{calinon2009statistical}, diffusion models~\cite{chi2023diffusion, urain2023se}, and bio-inspired techniques based on dynamical systems such as dynamical movement primitives (DMP)~\cite{hersch2008dynamical, pastor2009learning, saveriano2019merging, ijspeert2013dynamical}. However, these works do not attempt to intentionally encode the task constraints that are present in complex manipulation tasks. Furthermore, any path constraints, if encoded, depend on the coordinate frame attached to the end effector, i.e., the constraint representation is not coordinate-invariant. Thus, these algorithms can fail due to changes in the poses of the task-relevant objects, and there is no formal way to establish whether a generated plan is correct in terms of ensuring the task-relevant constraints are satisfied.

Recent work has proposed the use of screw-geometric representation of constraints (which are coordinate-invariant by construction) in complex manipulation tasks~\cite{laha2021point, mahalingam2023human}. It has been shown that the \emph{constraints, expressed as a sequence of constant screw motions, can be extracted from one  demonstration}~\cite{mahalingam2023human}. For new task instances, i.e., new poses of the task relevant objects, by suitably transferring the screw segments, and using screw linear interpolation (ScLERP)~\cite{sarker2020screw, mahalingam2023human} interleaved with Jacobian pseudo-inverse, one can generate joint-level motion plans that preserves path-based task constraints implicit in the demonstrations.  By focusing on the parts of the trajectory close to the objects of interest, these works pay particular attention to the constraints most relevant to the task. Therefore, they are much more robust to variations in the poses of the task relevant objects. However, they may fail if the Jacobian pseudo-inverse leads to a joint-space path that hits joint limits. Nevertheless, \emph{any feasible joint space plan (i.e., a plan satisfying the joint limits) generated using these approaches is guaranteed to satisfy the end-effector path constraints during motion}. Note that since, for tasks with end-effector constraints, it is much easier to computationally check joint space feasibility of paths compared to task space feasibility, we use the algorithm proposed in~\cite{mahalingam2023human} to plan motions. 

\subsection{Demonstration sufficiency using self-evaluation}

The high-level objective of our work is to develop a framework in which the robot can \textit{self-evaluate} its ability to generate low-level motion plans using a few demonstrations for complex manipulation tasks such as pouring, scooping, etc. This enables the robot to methodically seek additional demonstrations at targeted locations until it achieves a threshold of confidence in plan generation.

Previous works on self-evaluation of the satisfiability of task-related constraints~\cite{chatila2018toward, gautam2022method}, have focused mainly on sensor information available from a perception system~\cite{zillich2011knowing, kaipa2015toward} or high-level task specification~\cite{frasca2022framework}. However, none of the existing work on self-evaluation can be used to evaluate the robot's confidence in manipulation plan generation using kinesthetic demonstrations.

A recent work~\cite{trinh2024autonomous} explores \emph{demonstration sufficiency} in the context of navigation tasks through \textit{autonomous assessment}. While it focuses on demonstration phrased as expert-provided directions for navigating 2D grid environments, this approach does not directly extend to manipulation problems, where task constraints are embedded in the entire trajectory of the demonstration.


\section{Self-Evaluation}
\label{sec:solution-overview}


The \emph{pose of the end effector} refers to the position and orientation of the end effector frame -- the reference frame fixed to the link of the robot arm beyond the last joint or on any object that is rigidly held by the robot. We consider manipulation tasks where the task specification includes the initial pose of the end-effector and the poses of other task-relevant objects that determine the end-effector's goal pose.

\mypara{Task Instance}
Let $\mathcal{X} \subset SE(3)^n$ be a compact set, where $n$ is a positive integer representing the total number of task-relevant objects, including the object held by the robot. For a given manipulation task, an {\em instance of the task} is defined by ${\bf x} \in \mathcal{X}$. Thus, the set $\mathcal{X}$ defines the set of all task instances for a given task. We assume that all task-relevant objects are in the set of reachable poses of the robot's end effector.

\subsection{Motion Planning from a Single Demonstration and Evaluation of Motion Plan}

We will use the two-step motion planner discussed in~\cite{mahalingam2023human} to generate motion plans for new task instances that differ from the demonstration instance. In particular, we will denote this motion plan generator by $\texttt{hasMotionPlan}\left({\bf x},\ \bm{\Gamma}\right)$ which takes as input any task instance ${\bf x}$, a demonstration $\bm{\Gamma}$ and decides whether or not a motion plan can be generated for ${\bf x}$ using $\bm{\Gamma}$. The planning process involves \begin{enumerate*}[label=(\alph*)]
  \item transferring the guiding poses, $\bm{\Gamma}$, to a new set of guiding poses, $\bm{\Gamma}'$, to account for the new poses of the task-relevant objects and
  \item using screw linear interpolation (ScLERP) interleaved with Jacobian pseudo-inverse to generate the joint-space path between two consecutive guiding poses.
\end{enumerate*}
This motion planner ensures that the task space constraints in the demonstration represented as constant screw motion segments are always satisfied.  However, the motion planner may fail to generate a plan for a given task instance due to a violation of joint limits. Therefore, \emph{the planner used in this paper has the characteristic that when the planner returns a manipulation plan it is ensured that both task space and joint space constraints are satisfied}. Note that none of the other classes of planners for LfD, whether they be the more recent diffusion-based models or the classical DMP planners, can guarantee the satisfaction of path constraints in SE(3); making it difficult to formally (hence, computationally) characterize their performance.

Formally, we assume that given a demonstration $\bm{\Gamma}$ we have a mapping $f_{\bm{\Gamma}}: \mathcal{X} \rightarrow \{0, 1\}$, such that, for any new task instance ${\bf x} \in \mathcal{X}$, $f_{\bm{\Gamma}}({\bf x}) = 1$, if and only if the robot can successfully generate a manipulation plan for the task instance ${\bf x}$ using the demonstration $\bm{\Gamma}$. Note that the algorithmic implementation of $f_{\bm{\Gamma}}({\bf x})$ is $\texttt{hasMotionPlan}\left({\bf x},\ \bm{\Gamma}\right)$, defined above. Here, a plan is defined to be successful if it satisfies both task-space and joint-space constraints. If we consider a set of demonstrations $\mathcal{D} \equiv \sideset{}{_{i=1}^n}{\mathop{\{\bm{\Gamma}_i\}}}$ such that for any ${\bf x} \in \mathcal{X}$, $\displaystyle \max_{\bm{\Gamma}_i} \sideset{}{_{\bm{\Gamma}_i}}{\mathop{f}} ({\bf x}) = 1$, then we have a set of demonstration examples such that the robot is guaranteed to generate a manipulation plan for any instance of a given task.

We formally characterize the sufficiency of a set of demonstrations ${\mathcal{D}}$ to cover a given set of task instances $\mathcal{X}$. If $\mathcal{D}$ is considered insufficient by this metric, we describe an approach to identify a region (subset) of $\mathcal{X}$ that contains a good candidate for a new demonstration. This forms the basis for the incremental acquisition of the demonstrations further developed in \S\ref{sec:solution}.

\ignore{
We propose an incremental approach where the robot will use \emph{self-evaluation} to compute the set $\mathcal{E}$, incrementally, starting with an initial set of demonstrations $\mathcal{E}_0$, which can consist of only one demonstration. Self-evaluation is the process by which the robot evaluates the likelihood of successful plan generation for any arbitrary task instance, using a current set of demonstrations $\mathcal{E}_i$, and, in the process, identifies the region in its work area where plan generation is most likely to fail. 
}

\subsection{Sufficiency of Demonstrations}
For a demonstration, $\bm{\Gamma}_i$, let $\mathcal{B}(\bm{\Gamma}_i, \mathcal{X}) \subseteq \mathcal{X}$, be the set of task instances where $\bm{\Gamma}_i$ can be used to generate manipulation plans successfully. Thus,
\begin{align}
    \mathcal{B}\left(\bm{\Gamma}_i, \mathcal{X}\right) = \left\{ {\bf x} \in \mathcal{X} ~\middle| \sideset{}{_{\bm{\Gamma}_i}}{\mathop{f}} \left({\bf x}\right) = 1 \right\}
\end{align}
Let ${\rm Vol}(A)$, with $A \subseteq \mathcal{X}$ be a volume measure \cite{chirikjian2016harmonic} defined on $\mathcal{X}$. 

\begin{defn}[Coverage]
The \emph{coverage} of a demonstration $\bm{\Gamma}_i$
with respect to a set of task instances ${\cal X}$, denoted by $\mathbb{P}_\mathcal{X}(\bm{\Gamma}_i)$, is defined as:
\begin{align*}
    \mathbb{P}_\mathcal{X}(\bm{\Gamma}_i) = \frac{{\rm Vol}(\mathcal{B}(\bm{\Gamma}_i, \mathcal{X}))}{{\rm Vol} (\mathcal{X})}
\end{align*}
\end{defn}

\ignore{
This probability measure can also be defined on a subset of $\mathcal{X}$, say, \red{$\bar{\mathcal{X}}$}, as
\begin{align}
\label{eq:measure2}
    \mathbb{P}_{\bar{\mathcal{X}}}({\bf y}_i) = \frac{{\rm Vol}(\mathcal{B}({\bf y}_i) \cap \bar{\mathcal{X}})}{{\rm Vol} (\bar{\mathcal{X}})}
\end{align}
}

\ignore{
We will assume that $\mathcal{B}({\bf y}_i)$ is a union of a collection of {\em pairwise disjoint nonempty sets}, $\mathcal{B}_j({\bf y}_i) \subseteq \mathcal{X}$, such that ${\rm Vol}(\mathcal{B}_j({\bf y}_i)) \neq 0$. \red{This is a technical assumption that essentially disallows isolated single points to be part of the subset where the robot can successfully plan with a demonstration task instance ${\bf y}_i$.} 
}

Let $\mathcal{D} \equiv \sideset{}{_{i=1}^n}{\mathop{\{\bm{\Gamma}_i\}}}$ be a set of demonstrations. We can lift $\mathcal{B}$ from individual demonstrations to sets of demonstrations.  Let $\mathcal{B}(\mathcal{D}, \mathcal{X}) = \bigcup_{i=1}^{n} \mathcal{B}(\bm{\Gamma}_i, \mathcal{X})$. Then,
\begin{align}
\label{eq:Bdef}
    \mathcal{B}(\mathcal{D}, \mathcal{X}) = \left\{ {\bf x} \in \mathcal{X} ~\middle| ~\displaystyle\max_{{\bm{\Gamma}_i} \in \mathcal{D}} \sideset{}{_{\bm{\Gamma}_i}}{\mathop{f}} ({\bf x}) = 1 \right\}
\end{align}
Thus, $\mathcal{B}(\mathcal{D}, \mathcal{X})$ is the set of all task instances in $\mathcal{X}$ such that there is at least one demonstration $\mathcal{D}$ that can be used to generate a successful manipulation plan.  

Analogously, we can lift the notion of \emph{coverage} to sets of demonstrations:
\begin{align}
\label{eq:coverage}
    \mathbb{P}_\mathcal{X}(\mathcal{D}) = \frac{{\rm Vol}(\mathcal{B}(\mathcal{D}, \mathcal{X}))}{{\rm Vol} (\mathcal{X})}
\end{align}

\begin{defn}[Sufficiency]
Given a probability threshold $0 \leq \beta\leq 1$, a set of demonstrations $\mathcal{D}$ is said to be \emph{sufficient} for a set of task instances $\mathcal{X}$ with respect to $\beta$ if $\mathbb{P}_\mathcal{X}(\mathcal{D}) \geq \beta$
\end{defn}

\ignore{
For a given $\beta \in [0,1]$, {\em our goal is to obtain a set $\mathcal{E}$, of demonstration instances, incrementally, one at a time, such that}
\begin{align}
\label{eq:objective}
    \mathbb{P}_\mathcal{X}(\mathcal{E}) = \frac{{\rm Vol}(\mathcal{B}(\mathcal{E}))}{{\rm Vol} (\mathcal{X})} \geq \beta
\end{align}
{\em where $\mathbb{P}_\mathcal{X}(\mathcal{E})$ is the probability that the set of demonstration instances, $\mathcal{E}$, will lead to a successful plan generation for task instances chosen from the set $\mathcal{X}$}. 

The set of demonstrations corresponding to the task instances in $\mathcal{E}$ that satisfies~\eqref{eq:objective} is called a {\em sufficient set of demonstrations}. 
}

\ignore{
\begin{remark}
\st{There can be multiple variations of the above problem statement. For example, we can consider a variation in which we set a limit on the number of examples $m$, and our goal is to estimate $\beta$.}
\end{remark}

\begin{remark}
\st{One potential issue with the above problem formulation is that we are not considering the quality of a task demonstration explicitly. We are implicitly assuming that the setting is non-adversarial and the demonstrations provided are \textit{good enough}, where it is hard to quantify what is good enough?}
\end{remark}
}

\subsection{Identification of New Demonstration Candidates}
If a set of demonstrations $\mathcal{D}_0$ is insufficient for task instances $\mathcal{X}$, we would like to incrementally seek additional demonstrations, one at a time, thereby constructing a sequence of demonstration sets $\mathcal{D}_0,\mathcal{D}_1,\ldots$ such that some element of this sequence, say $\mathcal{D}_n$ is sufficient for $\mathcal{X}$.  

We identify a small subset of task instances to seek the next demonstration by partitioning $\mathcal{X}$ into $K$ disjoint compact sets $\mathcal{X}_j: 1 \leq j \leq K$.
Note that
\begin{align*}
    \mathbb{P}_\mathcal{X}(\mathcal{D}, \mathcal{X}) = \frac{{\rm Vol}(\mathcal{B}(\mathcal{D}, \mathcal{X}))}{{\rm Vol} (\mathcal{X})} = \frac{\sum_j {\rm Vol}(\mathcal{B}(\mathcal{D}, \mathcal{X}_j))}{\sum_j{\rm Vol} (\mathcal{X}_j)} 
\end{align*}
Then, if $\mathbb{P}_{\mathcal{X}}(\mathcal{D}_i) < \beta$, then there is a partition $\mathcal{X}_j$ such that $\mathbb{P}_{\mathcal{X}_j}(\mathcal{D}_i) < \beta$, that is, $\mathcal{D}_i$ is not sufficient for $\mathcal{X}_j$. Such a subset $\mathcal{X}_j$ contains candidate task instances for the next demonstration.

\subsection{Formulation as Multi-Arm Bandit Optimization}\label{sec:bandit-formulation}

We pose the problem of identifying the partition $\mathcal{X}_j$ that has the \emph{least coverage, 
$\mathbb{P}_{\mathcal{X}_j}(\mathcal{D}_i)$,} in terms of the $K$-arm bandit optimization problem~\cite{garivier2016optimal}. In the special case of Bernoulli $K$-arm bandit, each pull of the $j$-th arm yields a reward of $1$ with an (unknown) probability $p_j$ and $0$ with probability $(1-p_j)$.  The optimization problem is to find the \emph{best arm}, i.e., the one with the highest expected reward.  Algorithms for solving this problem are compared based on the expected number of pulls (samples) needed to determine the best arm.     
For our problem, we create a $K$-arm bandit with partitions $\mathcal{X}_j$ comprising the $K$ arms. Pulling the $j$-th arm corresponds to sampling a task instance ${\mathbf x}$ from $\mathcal{X}_j$. The reward $1$ if ${\mathbf x} \not\in \mathcal{B}(\mathcal{D}_i, \mathcal{X}_j)$, i.e. if the current set of demonstrations cannot generate a successful manipulation plan for ${\mathbf x}$; and $0$ otherwise. 

Note that in this formulation, the ``best arm'' corresponds to the partition that is least covered by $\mathcal{D}_i$.  In other words, the best arm points to the partition where the robot has the least ability and hence should ask for the next demonstration.
Once a new demonstration is obtained, the same process is repeated with the updated set of demonstrations $\mathcal{D}_{i+1}$. We stop when the current set of demonstrations is sufficient for all partitions.
\section{Incremental Demonstration Acquisition using Self-Evaluation}
\label{sec:solution}

\begin{algorithm}[t!]
    \caption{Obtaining a Sufficient Set of Demonstrations using \texttt{Self Evaluation}}\label{alg:self-evaluation}
    \hspace*{\algorithmicindent}\textbf{Input}: $K, \mathcal{X}_{1\ldots K}, \beta, \epsilon, \delta, \mathcal{D}_0$ \Comment{$\lvert \mathcal{D}_0\rvert \geq 1$}\\
    \hspace*{\algorithmicindent}\textbf{Output}: $\mathcal{D}$
    \begin{algorithmic}[1]

        \State $\mathcal{D} \gets \mathcal{D}_0$

        \While {true}
            \State $j^*, \hat{\mu}_{j^*}, \mathcal{T}_{j^*} \gets \texttt{getBestArm}\left(K, \epsilon, \delta, \mathcal{X}_{1\ldots K}, \mathcal{D}\right)$

            \If {$\hat{\mu}_{j^*} \leq 1 - \epsilon - \beta$} 
            \Comment{\small see \S\ref{sec:stopping-condition}}
                \State \textbf{break}
                \Comment{\small End demonstration acquisition}
            \EndIf    
            \State ${\bf y^*} \gets \texttt{selectFailedTaskIn}\left(\mathcal{T}_{j^*}\right)$ \Comment{see \S\ref{sec:suggest}}
            
            \State ${\bm\Gamma} \gets \texttt{getNewDemonstrationAt}\left(\bf y^*\right)$
            \Comment{see \S\ref{sec:obtain-demo}}
            
            \State $\mathcal{D} \gets \mathcal{D} \cup \{\bm\Gamma\}$

        \EndWhile
    
    \end{algorithmic}
\end{algorithm}

\begin{algorithm}[b!]
    \caption{\texttt{getBestArm}}\label{alg:get-best-arm}
    \hspace*{\algorithmicindent}\textbf{Input}: $K, \epsilon, \delta, \mathcal{X}_{1\ldots K}, \mathcal{D}$ \\
    \hspace*{\algorithmicindent}\textbf{Output}: $j^*, \hat{\mu}_{j^*}, \mathcal{T}^\mathbf{1}_{j^*}$
    \begin{algorithmic}[1]

        \State $N \gets \frac{2}{\epsilon^2}\ln\left(\frac{2K}{\delta}\right)$ 
        \Comment{\small see \S~\ref{sec:best-arm}}

        \For{$j \in \{1, \cdots, K\}$}

            \State $\mathcal{T}_j \gets \left\{ {\bf x} \sim \mathcal{U}({\mathcal{X}_j}) \right\}$ such that $\lvert \mathcal{T}_j \rvert = N$

            \State $\mathcal{T}^\mathbf{1}_j \gets \left\{{\bf x} \in \mathcal{T}_j \mid\ \lnot\texttt{hasMotionPlan}({\bf x},\ \bm{\Gamma})\quad \forall {\bm \Gamma} \in \mathcal{D}\right\}$ 

            \State $\hat{\mu}_j \gets \left\lvert \mathcal{T}^\mathbf{1}_j \right\rvert/\lvert \mathcal{T}_j \rvert$
            \Comment{\small Estimated expected reward}
        \EndFor

        \State $j^* \gets \displaystyle\arg\max_j \{\hat{\mu}_j\}$    
    \end{algorithmic}
\end{algorithm}

Algorithm~\ref{alg:self-evaluation} encodes our approach as pseudo-code to acquire demonstrations one at a time such that the set is sufficient for a given set of task instances $\mathcal{X}$.  The algorithm takes the $K$ partitions of task instances, $\mathcal{X}_{1\ldots K}$, the threshold $\beta$ to determine sufficiency, and two parameters $\epsilon$ and $\delta$, both $\in (0,1)$, and a non-empty set of initial demonstrations, $\mathcal{D}_0$. Each element of the set $\mathcal{D}_0$ is a sequence of guiding poses, where two consecutive guiding poses constitute a constant screw motion. The algorithm proceeds by adding new demonstrations, initialized as $\mathcal{D}_0$. In each iteration, 
\begin{enumerate}[label=(\alph*)]
\item The algorithm first determines the partition $j^*$ that is least covered by the current set of demonstrations (line 3) using bandit optimization as a subroutine (see \S\ref{sec:best-arm}). Along with $j^*$, we also obtain $\hat{\mu}_{j^*}$, the \emph{empirical} estimate of $1-\mathbb{P}_{\mathcal{X}_{j^*}}$ for the current set of demonstrations. 
\item These sample task instances are used to select a suggested task instance ${\bf y}^*$ for the next demonstration (line 4); see \S\ref{sec:suggest} for details.
\item Based on the suggested instance ${\bf y}^*$, we then obtain a new demonstration from the human teacher (line 5); see \S\ref{sec:obtain-demo} for details. 
\item The new demonstration is added to the set of current demonstrations (line 6). 
\end{enumerate}
The collection of new demonstrations ends when we determine, with high confidence, that the current set of demonstrations is sufficient for every partition (see \S\ref{sec:stopping-condition}).

\begin{figure}[!b]
    \centering
    \subfigure[Demonstration of \textbf{pouring}]{\resizebox{0.49\linewidth}{!}{
    \includegraphics[width=0.1633\linewidth]{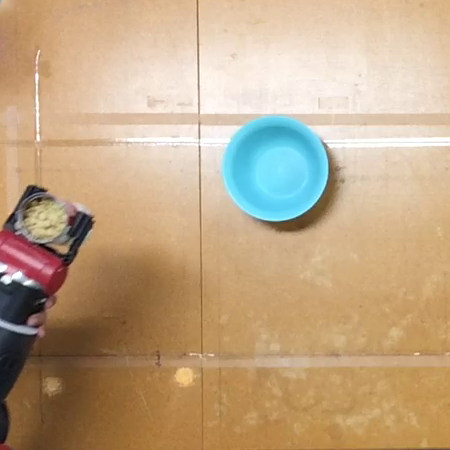}
    \includegraphics[width=0.1633\linewidth]{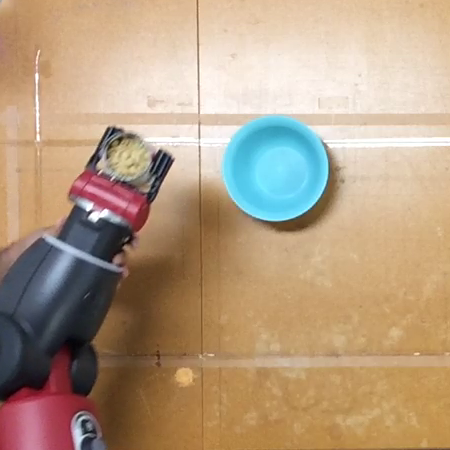}
    \includegraphics[width=0.1633\linewidth]{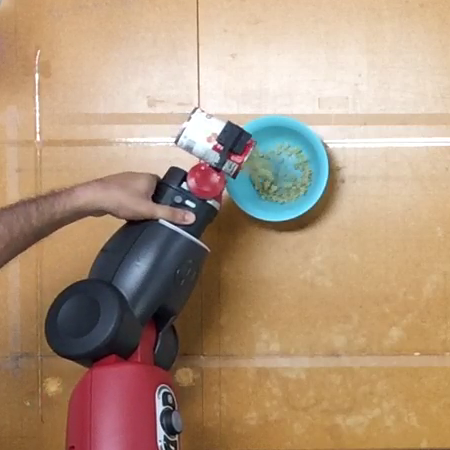}
  }\label{fig:pouring-demo}}
    \subfigure[Demonstration of \textbf{scooping}]{\resizebox{0.49\linewidth}{!}{
    \includegraphics[width=0.1633\linewidth]{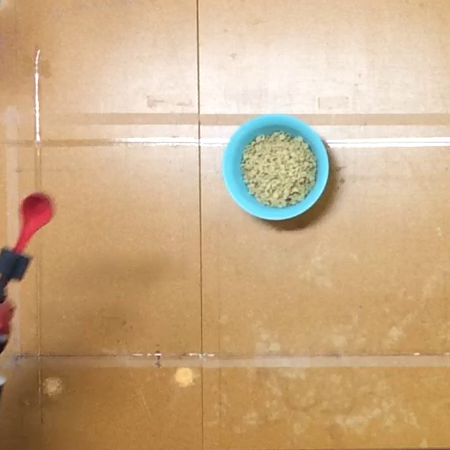}
    \includegraphics[width=0.1633\linewidth]{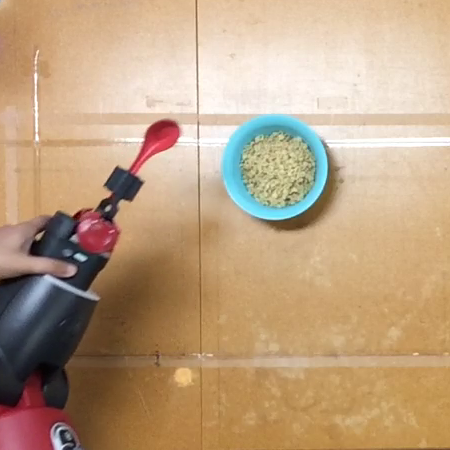}
    \includegraphics[width=0.1633\linewidth]{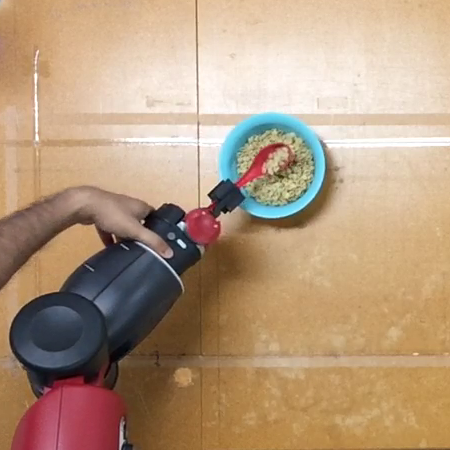}
  }\label{fig:scooping-demo}}
    \caption{Snapshots of kinesthetic demonstrations of \textbf{pouring} and \textbf{scooping} tasks. In each demonstration, the left frame is the initial pose and the right one is the final pose.}
    \label{Fig:demonstration}
\end{figure}

\begin{figure*}[!t]
    \centering
    \scalebox{0.9}{
        \resizebox{\textwidth + \columnsep}{!}{
            \begin{tikzpicture}[outer/.style={draw=gray,ultra thin,rounded corners,densely dashed, fill=gray!5,inner xsep=0ex,inner ysep=0pt,yshift=0ex,fit=#1}]

                    \node[label={[yshift=6pt]below:\scalebox{0.4}{\begin{tabular}{@{}c@{}}
                       \textbf{[1]} Object Location for the\\ \emph{Initial} Demonstration
                     \end{tabular}}}] (A) at (-0.5,1.5) {\includegraphics[width=1.8cm]{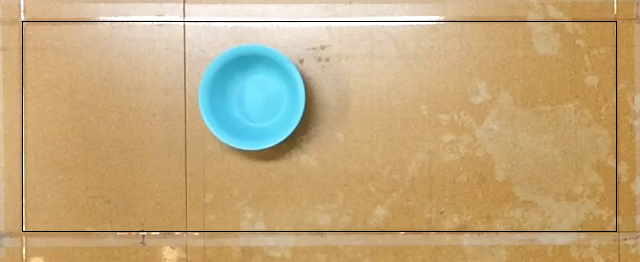}};

                    \node[label={[yshift=6pt]below:\scalebox{0.4}{\textbf{[2]} Estimated Object Pose in Robot's Work Area}}] (B) at (1,-0.2) {\includegraphics[width=3cm]{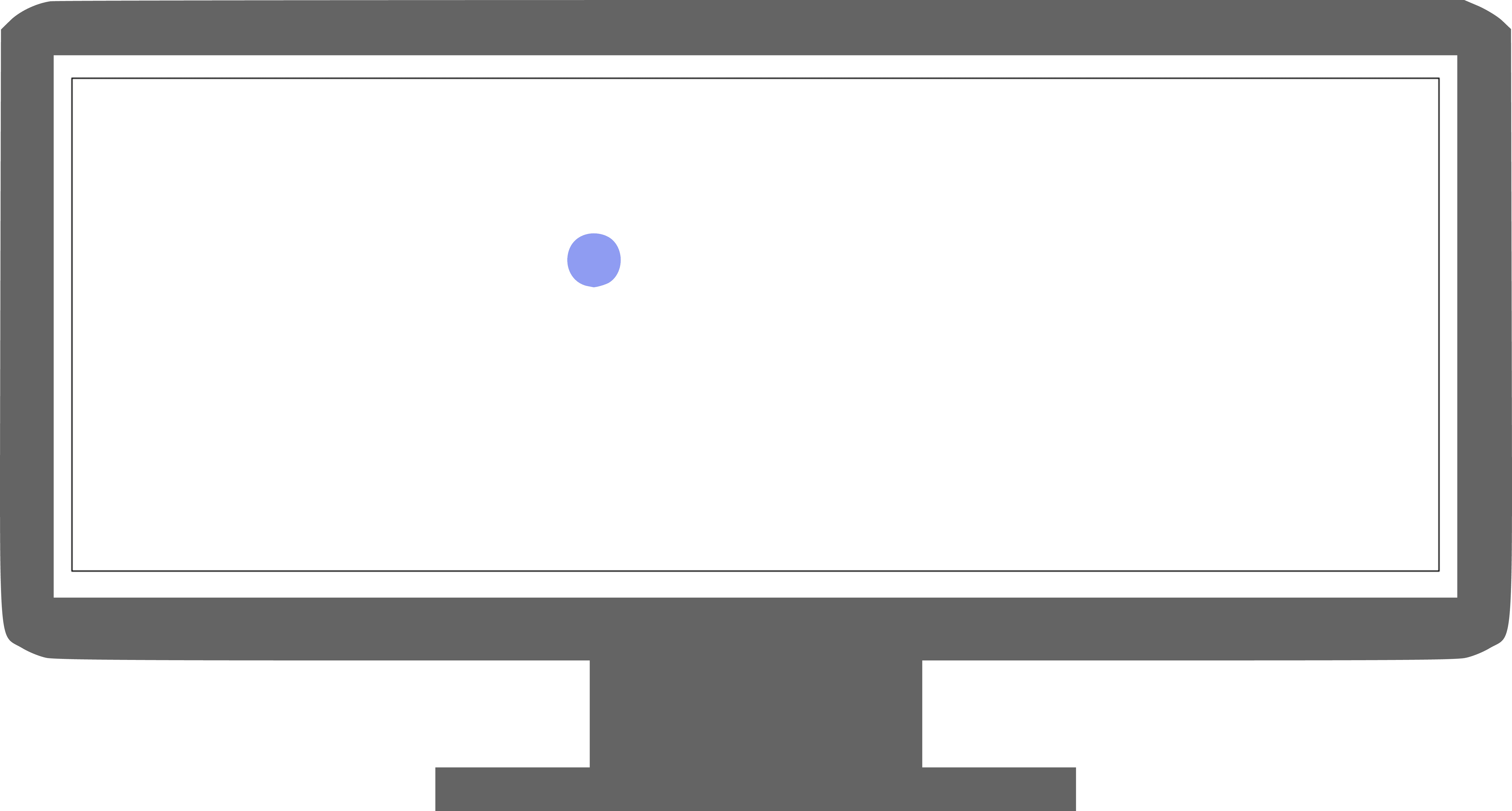}};

                    \draw[white,thick,dotted,line width=0.5pt,line cap=round,decoration={markings,mark=at position 0.6 with {\arrow{stealth'[darkgray!80]}}},postaction={decorate}] (A.south east) to ([xshift=-5pt]B.north);

                    \node[label={[yshift=6pt]below:\scalebox{0.4}{\textbf{[3]} Demonstration Acquisition}}] (C) at (2.8,1.5) {\includegraphics[width=1.8cm]{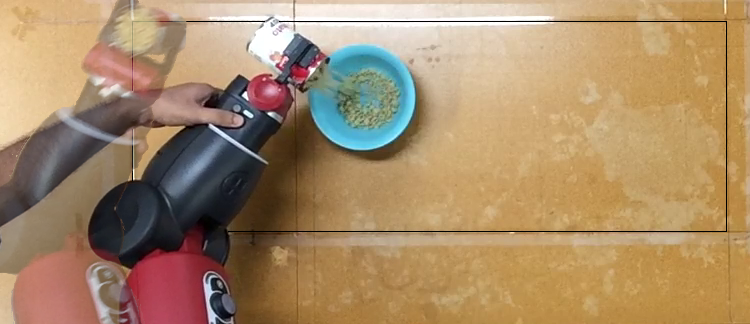}};

                    \draw[white,thick,dotted,line width=0.5pt,line cap=round,decoration={markings,mark=at position 0.8 with {\arrow{stealth'[darkgray!80]}}},postaction={decorate}] ([xshift=15pt]B.north) to (C.south west);

                    \node[label={[yshift=6pt]below:\scalebox{0.4}{\textbf{[4]} Robot's Belief after the first Demonstration}}] (D) at (4.5,-0.2) {\includegraphics[width=3cm]{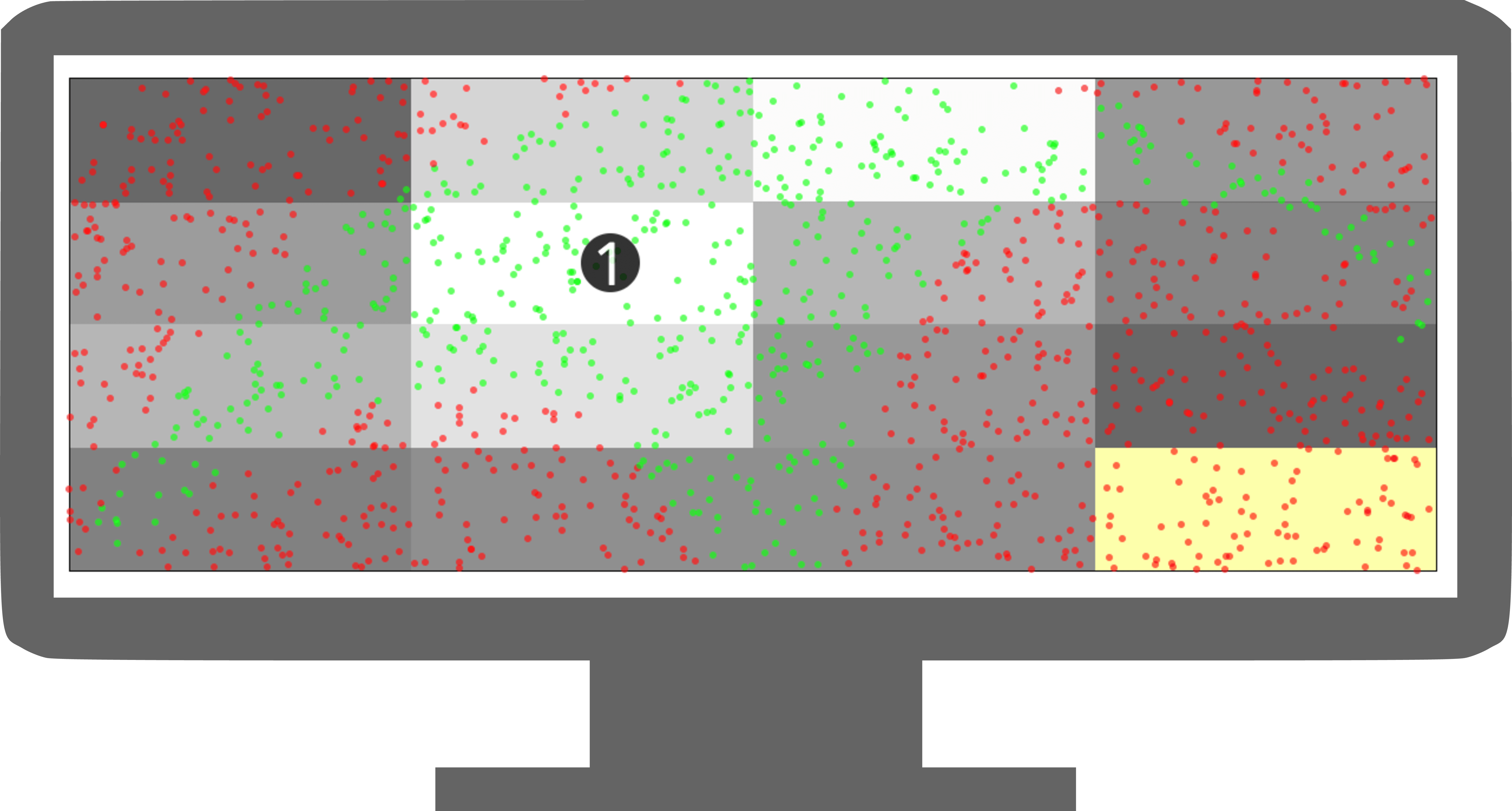}};

                    \draw[white,thick,dotted,line width=0.5pt,line cap=round,decoration={markings,mark=at position 0.6 with {\arrow{stealth'[darkgray!80]}}},postaction={decorate}] (C.south east) to ([xshift=-10pt]D.north);

                    \node[label={[yshift=6pt]below:\scalebox{0.4}{\begin{tabular}{@{}c@{}}
                       \textbf{[5]} Recommended Location\\ for the \emph{Next} Demonstration
                     \end{tabular}}}] (E) at (6.2,1.5) {\includegraphics[width=1.8cm]{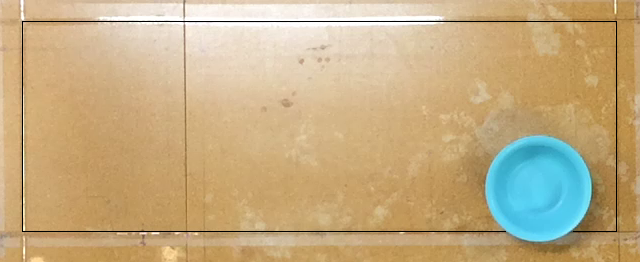}};

                     \draw[white,thick,dotted,line width=0.5pt,line cap=round,decoration={markings,mark=at position 0.7 with {\arrow{stealth'[darkgray!80]}}},postaction={decorate}] ([xshift=10pt]D.north) to (E.south west);

                     \node[label={[yshift=6pt]below:\scalebox{0.4}{\textbf{[6]} Estimated Object Pose for the next Demonstration}}] (F) at (8,-0.2) {\includegraphics[width=3cm]{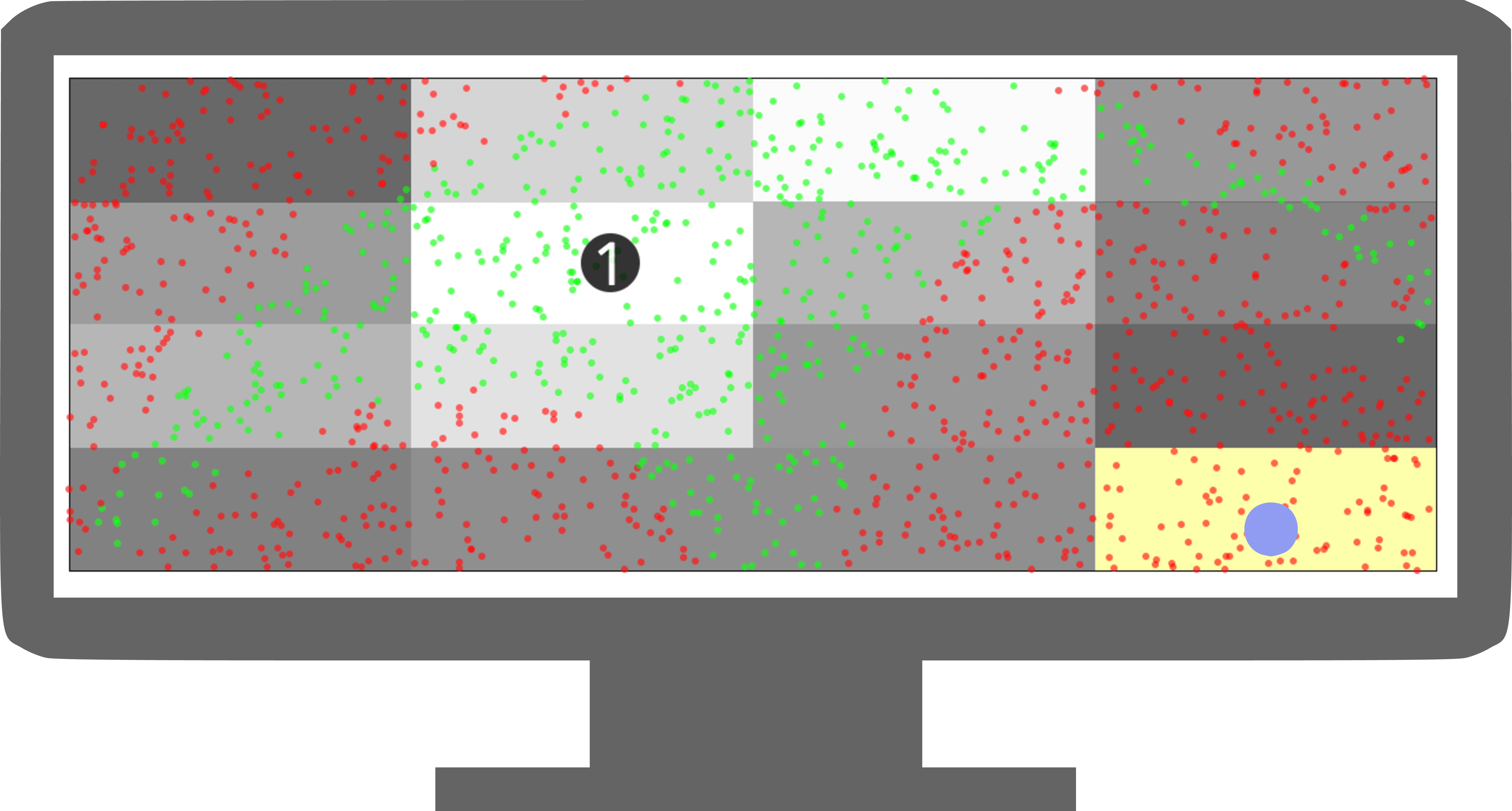}};

                     \draw[white,thick,dotted,line width=0.5pt,line cap=round,decoration={markings,mark=at position 0.6 with {\arrow{stealth'[darkgray!80]}}},postaction={decorate}] (E.south east) to ([xshift=-10pt]F.north);
                     \node[xshift=-2pt,yshift=2pt] (G) at (9,1.2) {$\cdots$};

                     \draw[white,thick,dotted,line width=0.5pt,line cap=round,decoration={markings,mark=at position 0.5 with {\arrow{stealth'[darkgray!80]}}},postaction={decorate}] ([xshift=10pt]F.north) to (G.south west);
            \end{tikzpicture}
        }
    }
    \caption{Workflow of the demonstration acquisition process guided by the \emph{self evaluation} algorithm. The upper row shows the events happening in the physical environment while the lower row shows the corresponding results on a user interface with the blue dot representing the estimated object pose.}
    \label{fig:overview}
    \vspace{-\baselineskip}
\end{figure*}

\subsection{Finding an $\epsilon$-Optimal Arm}
\label{sec:best-arm}
Algorithm~\ref{alg:get-best-arm} uses the na\"ive $(\epsilon, \delta)$-PAC learning algorithm to identify an $\epsilon$-optimal arm of our bandit formulation. In this algorithm, $\mathcal{T}_j^{\mathbf{1}}$ (line 4) is the set of all task instances for which we cannot find a motion plan and hence have reward 1.  This algorithm ensures with confidence $1 - \delta$ that the empirical estimate of the expected reward for each arm, $\hat{\mu}_j$, is $\epsilon$-close to the true expected reward $\mu_j$.
That is, \begin{small}$\displaystyle \mathbb{P}\left(\max_j\left\{\left\lvert \mu_j - \hat{\mu}_j \right\rvert\right\} \le \epsilon\right) \ge 1 - \delta$\end{small}. 

The samples for each arm are drawn from uniformly distributed random task instances in each partition.  Since each partition is a subset of $SE(3)$, we can use the techniques described in \cite{kuffner2004effective} used for sampling from $SE(3)$. The reward for each sample ${\bf x}$ is determined by checking if an executable motion plan can be derived for ${\bf x}$ based on the currently available set of demonstrations. We use the ScLERP based motion planner~\cite{mahalingam2023human} to generate plans for each sample task instance. The ScLERP motion planner generates a set of guiding poses for each given task instance, but may fail to produce a sequence of feasible joint configurations needed to execute the plan even when such feasible configurations exist. This is a fundamental problem due to the inherent difficulty in always finding feasible paths in the joint space via inverse kinematics. Hence, determining rewards via ScLERP motion planner gives us an upper bound of rewards.  Nevertheless, this only means that the set of demonstrations we find will be \emph{conservative} --- sufficient to cover the task instances --- but may not be minimal. The minimum number of samples drawn from each arm, $N$, needed to ensure the confidence bound, is $\frac{2}{\epsilon^2}\ln\left(\frac{2K}{\delta}\right)$.  To identify $j^*$ such that $\mu_{j^*}$ is within $\epsilon$ of the actual maximum, we need to estimate the probabilities of the regions to within $\frac{\epsilon}{2}$ of their actuals in the worst case.  When used with Hoeffding's inequality~\cite{Hoeffding01031963}, this gives rise to the $\frac{2}{\epsilon^2}$ factor in the bound for $N$. 

Note that we do not need to estimate each $\mu_j$ accurately to identify $j^*$; only the expected rewards of arms that are close to optimal need to be evaluated accurately. Several PAC-learning algorithms reduce the number of samples needed to identify an arm that is $\epsilon$-close to optimal with confidence $(1-\delta)$~\cite{mannor2004sample}. Other techniques such as UCB~\cite{auer2002finite} may also be used to identify $j^*$. We are currently evaluating the suitability of such techniques and their effect on incremental demonstration acquisition.

\begin{figure*}[!ht]
    \vspace{\baselineskip}
    \centering
    \begin{tikzpicture}
        \newdimen\R
        \newdimen\W
        \R=0.5\linewidth
        \W=0.31\linewidth
        \node (1) at (180:\R) {\includegraphics[width=0.95\W]{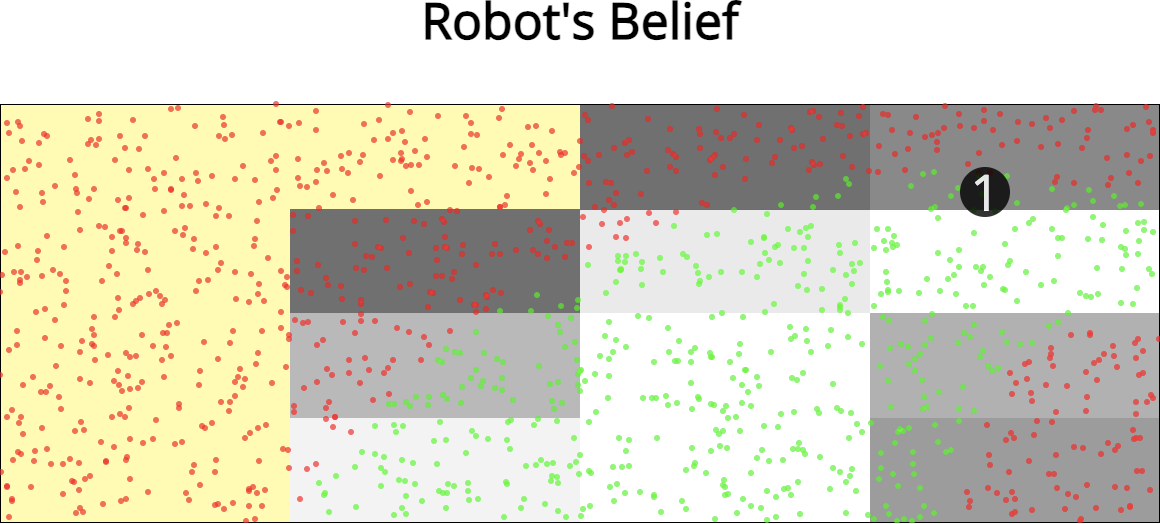}};
        \node (2) at (0:-0.5\W) {\includegraphics[width=0.95\W]{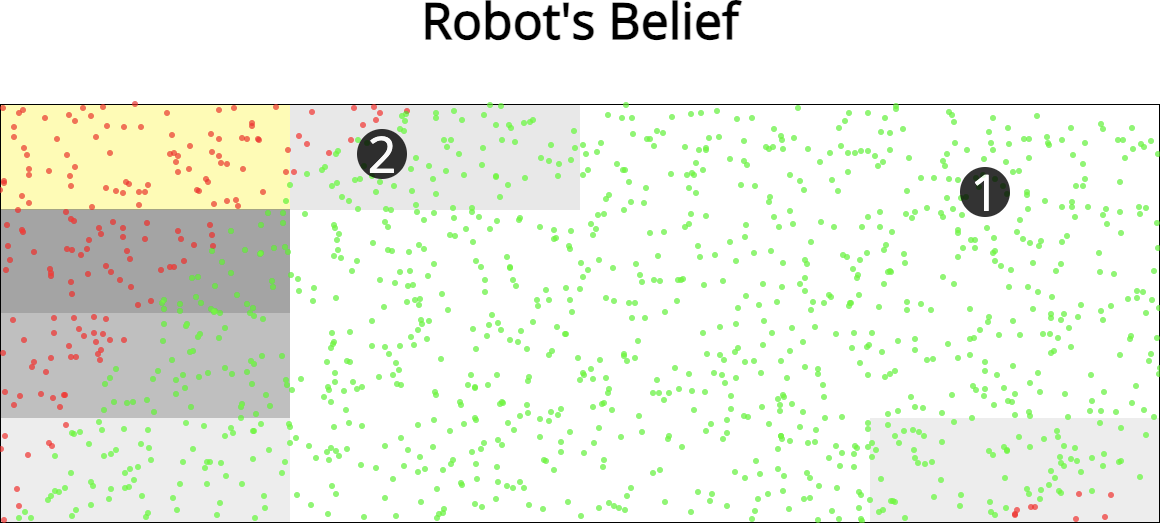}};
        \node (3) at (0:\R-\W) {\includegraphics[width=0.95\W]{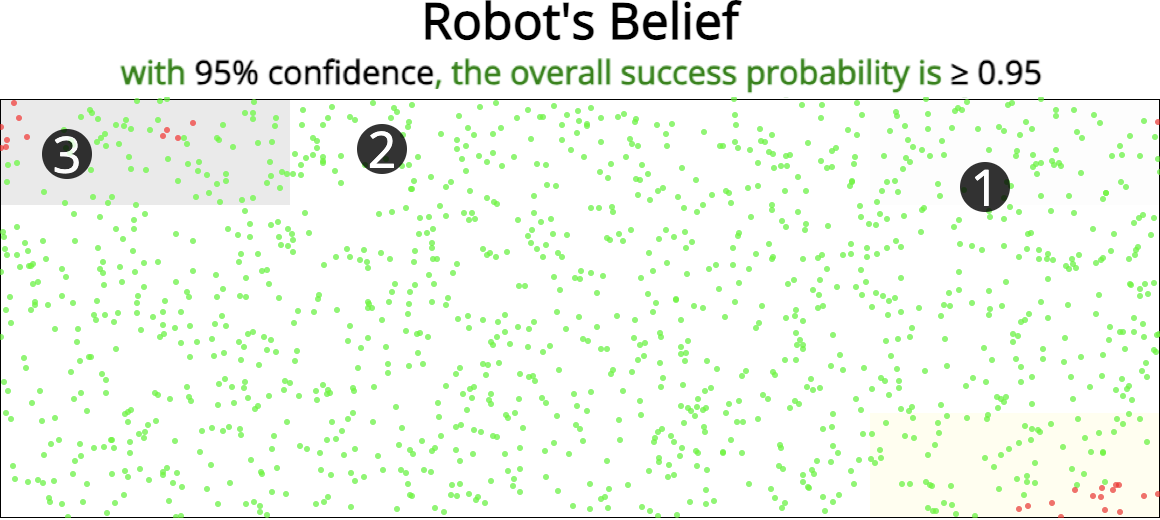}};
        
        \node [rectangle callout,rounded corners,fill=assisted!50,opacity=0.6,overlay,callout absolute pointer={([xshift=-28pt,yshift=-27pt]1.north east)}] at ([yshift=5pt]1.north) {\footnotesize Object Location for Demonstration \#1};

        \node [rectangle callout,rounded corners,fill=assisted!50,opacity=0.6,overlay,callout absolute pointer={([xshift=53pt,yshift=-22pt]2.north west)}] at ([yshift=5pt]2.north) {\footnotesize Object Location for Demonstration \#2};

        \node [rectangle callout,rounded corners,fill=assisted!50,opacity=0.6,overlay,callout absolute pointer={([xshift=14pt,yshift=-23pt]3.north west)}] at ([yshift=5pt]3.north) {\footnotesize Object Location for Demonstration \#3};
        
        \draw[white,thick,line width=2pt,decoration={markings,mark=at position 0.8 with {\arrow{stealth'[gray]}}},postaction={decorate}] ([yshift=-10pt]1.east) to ([yshift=-10pt]2.west);
        \draw[white,thick,line width=2pt,decoration={markings,mark=at position 0.8 with {\arrow{stealth'[gray]}}},postaction={decorate}] ([yshift=-10pt]2.east) to ([yshift=-10pt]3.west);
    \end{tikzpicture}
    \caption{Evolution of the robot's belief (shown as probability heat-map in varying shades of gray with white representing the success probability of $1.0$) for one participant providing demonstrations of scooping using the guidance of \emph{self evaluation} (Algorithm~\ref{alg:self-evaluation}). Green and red dots represent locations of the task instances where plan generation succeeded and failed respectively. Yellow region is the recommendation provided by Algorithm~\ref{alg:self-evaluation} for obtaining the next demonstration.}
    \label{fig:heatmap}
    \vspace{-\baselineskip}
\end{figure*}

\subsection{Suggestion for the Next Demonstration}
\label{sec:suggest}
Having determined the partition with the least coverage, we seek the next demonstration using the following heuristic.  
\ignore{
In lines $4$ and $5$ of Alg.~\ref{alg:self-evaluation}, we obtain a new kinesthetic demonstration ${\bm \Gamma}_{y}\left({\bm \Theta}_{y}\right)$ from the region $\mathcal{X}_{j^*}$ (corresponding to ${\bf y^*}$), by first selecting a failed task instance ${\bf y^*}$ from the corresponding set of failed task instances $T_{j^*}$, using a heuristic on the screw-segment where the task instance failed.}
Since we use ScLERP-based motion planner to check for feasible motion plans, we know that for each failed task instance (i.e. those in $T_{j^*}$ with reward 1), there will be a screw segment $\left({\bf g}^{(i)}, {\bf g}^{(i+1)}\right) \in \mathcal{G}$ such that ${\bf g}^{(i)}, {\bf g}^{(i+1)} \in \mathcal{G}$ for which we could not find a feasible joint-space path (due to joint limit violations).  We choose the failed task instance ${\bf y^*}$ for which the joint-limit violation occurred in the earliest screw segment. Intuitively, a task failing early may have fewer viable alternative executions.

\subsection{Obtaining a New Demonstration}
\label{sec:obtain-demo}
A kinesthetic demonstration is collected, by recording a discrete sequence of joint angles $\mathbf{\Theta}$ and extracting from them a sequence of guiding poses $\mathbf{\Gamma}$~\cite{mahalingam2023human}. Note that even though we were seeking a demonstration for task instance $\mathbf{y}^*$, the task instance for the provided demonstration may be different -- object poses may differ due to errors in manual placement.

\subsection{Stopping Condition}
\label{sec:stopping-condition}
This incremental process continues until the robot is confident enough that the overall success probability in the entire work area meets or exceeds a chosen threshold $\beta$ i.e., $\mathbb{P}_{\mathcal{X}}(\mathcal{D}) \geq \beta$.
In our setup $\mu_{j^*}$ is the (true) probability of failure to generate a successful motion plan in the region $\mathcal{X}_{j^*}$ corresponding to the optimal bandit arm $j^*$. When $\mu_{j^*} \leq 1 - \beta$, we can then claim with confidence $1 - \delta$ that for each partition $\mathcal{X}_j$, $\mathbb{P}_{\mathcal{X}_j}(\mathcal{D}_i) \geq \beta$.  Note that we only know the empirical estimate $\hat{\mu}_j$ with the constraint
$\lvert \mu_j - \hat{\mu}_j\rvert \leq \epsilon$.  The \emph{testable} condition used as the stopping condition (line 4 of Alg.~\ref{alg:self-evaluation}) is $\hat{\mu}_{j^*} \leq 1 -\epsilon - \beta$ since it implies $\mu_{j^*} \leq 1 - \beta$.
When this condition is satisfied, we can claim with confidence $(1-\delta)$ that the current set of demonstrations $\mathcal{D}$ is sufficient for all task instances in $\mathcal{X}$.

Note that although $\mu_{j^*}$ is monotonically non-increasing, we cannot bound the number of demonstrations needed to satisfy the stopping condition {\em a priori}. Alternatively, if we use a budget of a maximum number of demonstrations as our stopping criterion or stop early, we can compute the probability $\beta' = 1-\epsilon-\hat{\mu}_{j^*}$ such that for each partition the robot believes it will be successful with probability $\beta'$. Here, $\hat{\mu}_{j^*}$ is the largest empirically estimated failure probability among all regions in the last iteration.
\section{User Study Experiments and Results}
\label{sec:user-study}

\begin{figure}[!b]
    \resizebox{0.9\linewidth}{!}{
        \includegraphics[]{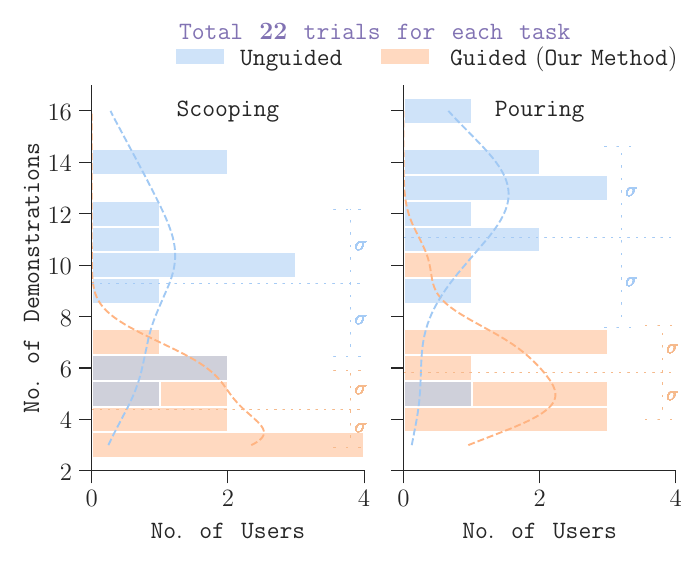}
    }
    \caption{Distribution of the number of demonstrations required to achieve the same objective of at least $95\%$ overall success-rate with at least $95\%$ confidence in Guided vs. Unguided trials.}
    \label{fig:demonstration_distribution}
\end{figure}

To assess the effectiveness of incremental demonstration acquisition, we performed a systematic user study acquiring demonstrations in two kinds of trial: \textbf{\emph{guided}} and \textbf{\emph{unguided}}. In a \emph{guided} trial, an interface showed the positions of task-relevant objects used in previous demonstrations (acquired using an object detection and localization system), a heatmap showing task instances for which motion planning using the given demonstrations succeeded or failed, and specifically identified candidate regions for the next demonstration (see \S\ref{sec:suggest}).  In an unguided trial the users were only shown the positions of task-relevant objects in previous demonstrations, but not the robot's belief or a candidate for next demonstration. We used two tasks: \textbf{\emph{pouring}} and \textbf{\emph{scooping}} as examples of \textit{complex manipulation tasks} (see Fig.~\ref{Fig:demonstration}). These tasks were chosen since they are characterized by constraints on the end-effector's motion that are key for their successful execution but difficult to specify beforehand, especially for non-experts. Moreover, the motion constraints in these two tasks differ significantly, making them well suited for the user study. Performing one after the other does not introduce any learning biases to the participants.

\subsection{User Experiment Setup and Participants}
\label{sec:experimental-setup}

\begin{table}[!b]
    \centering
    \begin{tabular}{c|c|c|c|c}
      \toprule
      \multirow{4}{*}{Task} & \multicolumn{4}{c}{Number\ \ of\ \ Sufficient\ \ Demonstrations} \\
      \cmidrule{2-5} & \multicolumn{2}{c}{\cellcolor{unassisted!50}Unguided} & \multicolumn{2}{|c}{\cellcolor{assisted!50}Guided (Our Method)} \\
      \cmidrule{2-5} & mean ($\mu$) & variance ($\sigma^2$) & mean ($\mu$) & variance ($\sigma^2$) \\
      \midrule
      Scooping & 9.7 & 9.42 & 4.5 & 2.07 \\
      Pouring & 11.9 & 8.69 & 5.8 & 3.36 \\
      \bottomrule
    \end{tabular}
    \caption{Number of \emph{sufficient kinesthetic demonstrations} required to achieve a pre-determined success-rate of at least $95\%$ for two complex manipulation tasks over 22 trials each.}
    \label{tab:demonstrations_count}
\end{table}

The experiments were carried out in a tabletop environment using the Baxter robot from ReThink Robotics using a rectangular work area of $0.38 m \times 1.055 m$, which is a significant fraction of the reachable area of the robot on the table (see Fig.~\ref{Fig:problem}). We used the Intel RealSense D415 RGB-D depth camera for vision sensing. The object's pose is defined as that of the 3D bounding box enclosing its corresponding 3D point cloud. Object detection is achieved using YOLOv5~\cite{jocher2022ultralytics}, with detected pixels projected back to $\mathbb{R}^3$ using intrinsic camera parameters and depth values. Off-the-shelf algorithms are used for point-cloud processing and 3D bounding-box computation.

For the experiments, we chose $K=16,\epsilon=0.02,\beta=0.95,$ and $\delta=0.05$ for Alg.~\ref{alg:self-evaluation}. Thus, the objective of each participant was to provide a sufficient number of kinesthetic demonstrations for each task, so that, with $95\%\ (\text{i.e. }1-\delta)$ confidence, the robot could generate manipulation plans for at least $95\%\ (\text{i.e. }\beta)$ of the task instances uniformly sampled from the work area. There was no limit on the amount of time or the number of demonstrations for each trial, i.e., each trial continued until the objective was achieved.

The study was carried out with $22$ adult graduate students consisting of $5$ female and $17$ male, none of them having prior experience interacting with robots. The study lasted about $1.5$ hours on average for each participant. Each participant underwent an initial training phase in which they were instructed about the tasks. They also moved the arm around in zero-gravity mode and the study commenced once the participants self-reported that they were comfortable in moving the arm anywhere in the workspace.

\subsection{Experiment Design}
\label{sec:experiment-design}

Two types of trials (\emph{guided} and \emph{unguided}) applied to two tasks (\emph{pouring} and {\emph{scooping}) lead to four experimental conditions: \emph{guided pouring}, \emph{unguided pouring}, \emph{guided scooping}, and \emph{unguided scooping}. To ensure that the trials were completed in a reasonable time, each participant was assigned to two different experimental conditions, always involving different tasks, while maintaining a fixed order: \emph{the unguided condition always preceding the guided condition} -- either (unguided pouring, guided scooping) or (unguided scooping, guided pouring). This design helps mitigate any potential bias introduced by the task order.

\subsection{Experimental Findings}
\label{sec:discussion}

Since each of the $22$ users was asked to participate in $2$ (different tasks) of the $4$ experimental conditions, for each experimental condition, we had a total of $11$ trials. We now present our key experimental results.

\subsubsection{Number of Kinesthetic Demonstrations} For both guided and unguided trials, each participant was able to teach the robot to achieve the given overall planning success rate of at least $95\%$ with high confidence (at least $95\%$). The fact that each participant could teach the robot in guided and unguided trials provides empirical validation of the choice of the ScLERP-based planning approach. However, as shown in Table~\ref{tab:demonstrations_count} and Fig.~\ref{fig:demonstration_distribution}, for the unguided trials, the participants on average had to give more demonstrations. Using Alg.~\ref{alg:self-evaluation} helped reduce the number of demonstrations by $53.6\%$ and $51.3\%$ for the scooping and pouring tasks, respectively. Only $5$ to $6$ demonstrations were needed on average. Furthermore, as Table~\ref{tab:demonstrations_count} shows, the variability between the participants was much smaller in the guided trial than in the unguided trial. Thus, \emph{self evaluation} also ensures more uniformity in the teaching outcome of the participants despite potential individual differences in their ability to provide demonstrations.

Fig.~\ref{fig:evolution_of_success_probability} shows the variation in the overall success probability with the number of demonstrations. In unguided trials, although initially the success probability increased quickly, it soon plateaued, requiring significantly more demonstrations to reach the desired threshold of $0.95$. Thus, without guidance, participants struggled to identify the remaining failure-prone areas in the workspace. However, with guidance, such a pronounced plateauing effect was not present.

\begin{figure}[!ht]
    \vspace{-\baselineskip}
    \centering
    \resizebox{\linewidth}{!}{
        \includegraphics[]{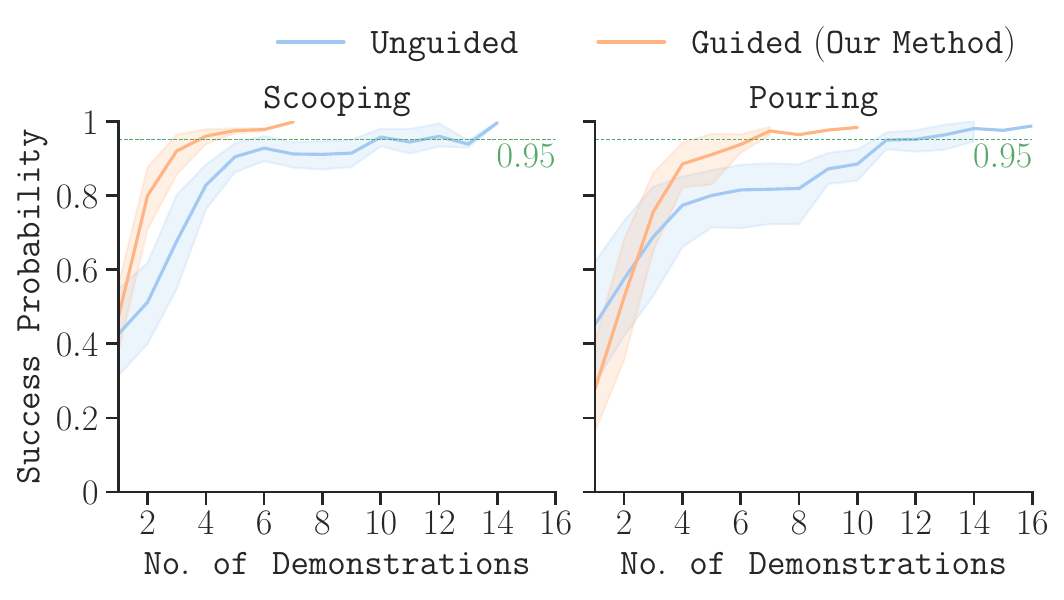}
    }
    \caption{Change of the Overall Success Probability with Incremental Acquisition of Kinesthetic Demonstrations. Thick lines represent the mean of all the 11 trials shown as the bands.}
    \label{fig:evolution_of_success_probability}
    \vspace{-0.25\baselineskip}
\end{figure}

\section{Simulation Experiments and Results}
\label{sec:simulation-results}

The number of demonstrations obtained in a single trial in the user study may vary with different executions and also depends on the choice of $K$. To get an idea of the distribution of the number of sufficient demonstrations, we performed $1000$ simulation trials, using self-evaluation, each for a different value of the number of disjoint regions $K$. The other parameters: $\epsilon$, $\delta$, and $\beta$ were kept the same in all executions as before. Note that Alg.~\ref{alg:self-evaluation} is an interactive process in which after each iteration the robot asks for the next demonstration from a human until it achieves its desired confidence. To bypass the human-robot interaction after each iteration in this simulation experiment, we pre-collected $32$ demonstrations with at least one demonstration in any selected region. One of these pre-collected demonstrations was chosen in each iteration. The result of the simulation is shown in Fig.~\ref{Fig:demo_distribution}. The distributions show that we required a maximum of $7$ and $6$ demonstrations for pouring and scooping, respectively. Furthermore, in every simulation, for each value of $K$, the estimated expected number of sufficient demonstrations is always within $\pm 0.08$ of the $99\%$ confidence interval; and as $K$ increases, the expected number of demonstrations increases only marginally (for example, for $K = 16$, it required only $3$ to $4$ demonstrations, on average, for both tasks). Note that smaller values of $K$ require a smaller number of samples (from the bandit formulation). However, with smaller values of $K$, although we may achieve the desired confidence in the entire work area, there may be regions with much lower local success probabilities. This effect is explored in detail below.

\begin{figure}[!t]
    \centering
    \resizebox{\linewidth}{!}{
        \includegraphics[]{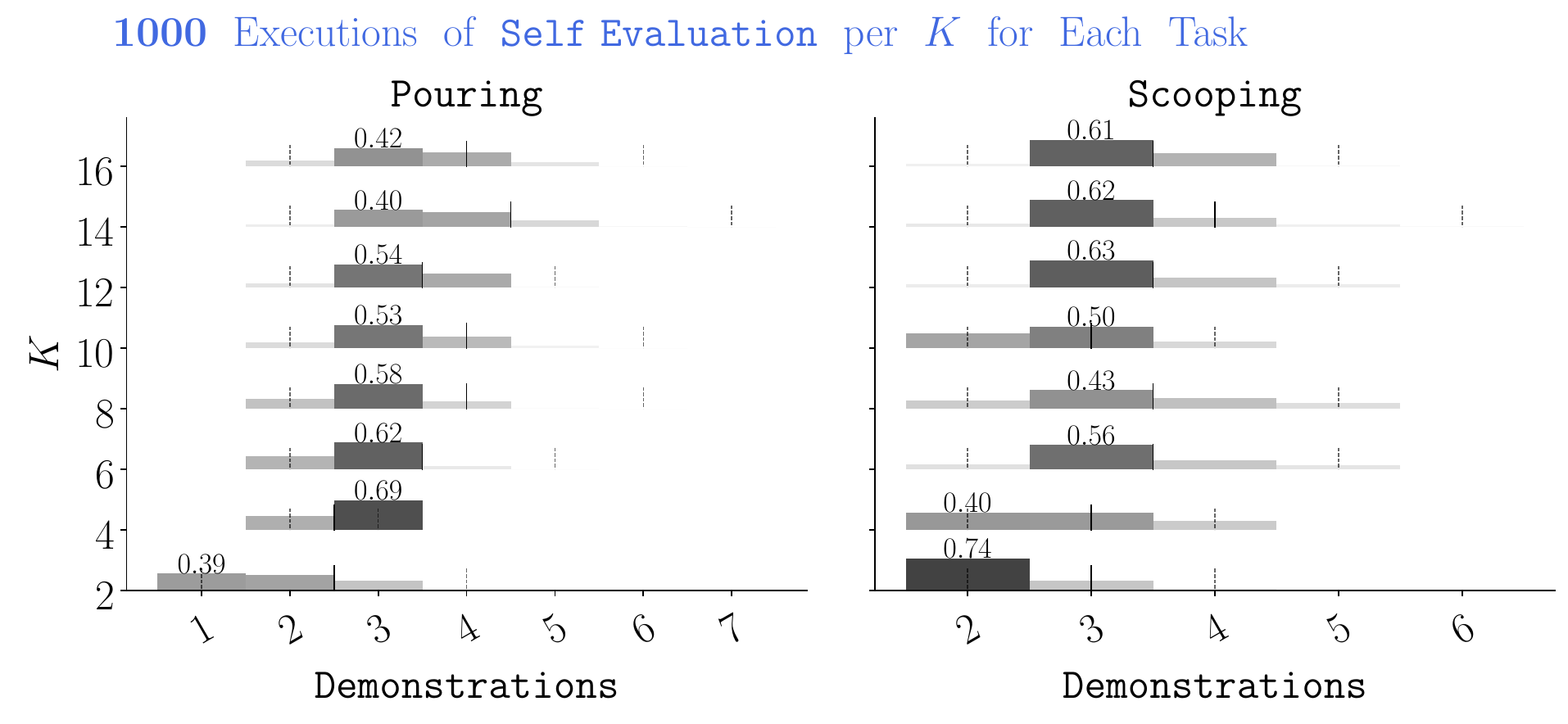}
    }
    \caption{Distribution (probability mass function) of the number of sufficient demonstrations for different values of the number of regions $K$. For each $K$, Algorithm~\ref{alg:self-evaluation} was executed $1000$ times. The $\min$, mean, mode and $\max$ are highlighted for each distribution.}
    \label{Fig:demo_distribution}
\end{figure}

\begin{figure}[!b]
    \vspace{-0.5\baselineskip}
    \centering
    \subfigure[$K=1$]{\resizebox{0.48\linewidth}{!}{
        \includegraphics[]{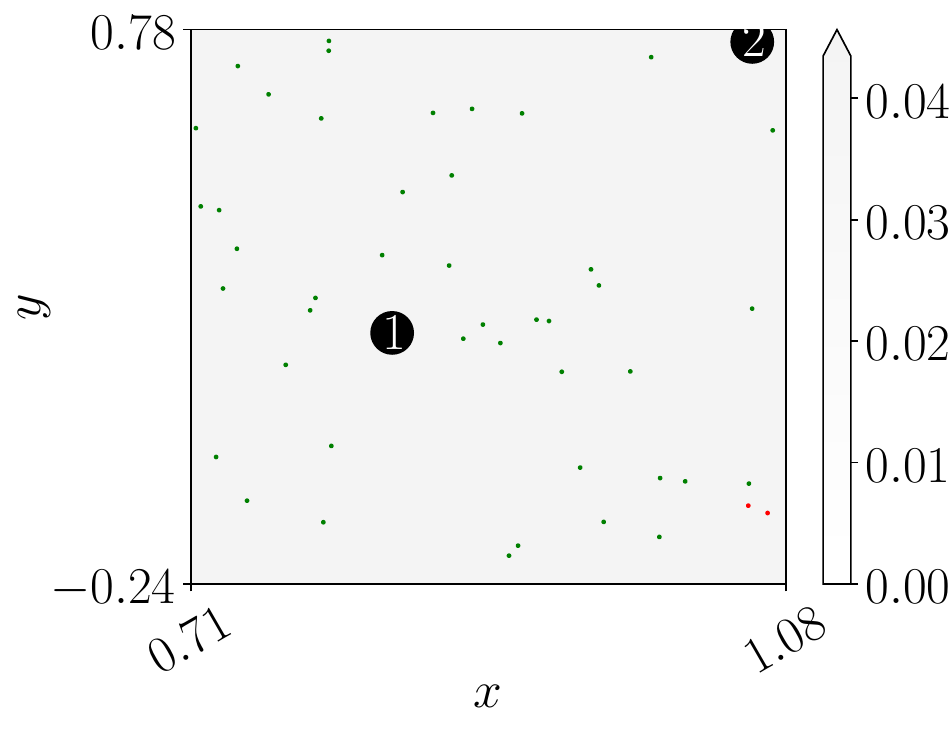}
    }}
    \subfigure[$K=16$]{\resizebox{0.47\linewidth}{!}{
        \includegraphics[]{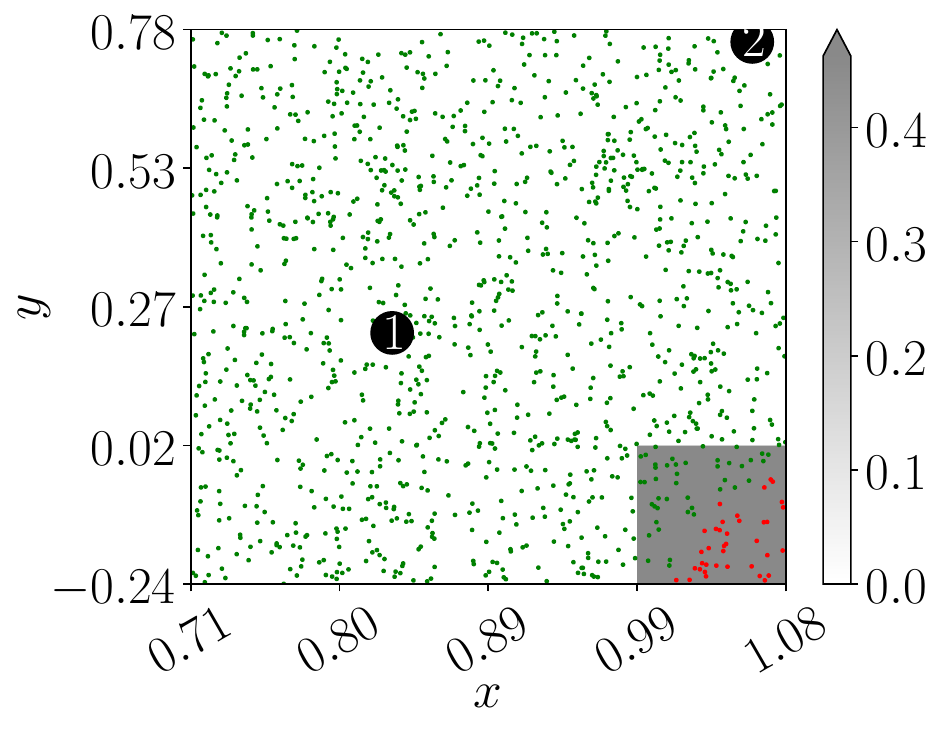}
    }}
    \caption{Effect of choosing two different values of the number of disjoint regions $K$, but with the same demonstrations.}
    \label{Fig:justification-for-K}
\end{figure}

For smaller values of $K$, the lower number of demonstrations (see Fig.~\ref{Fig:demo_distribution}) is due to the fact that we had fewer task samples to evaluate and the success guarantee is also not fine-grained, in the sense that although the overall probability guarantee may be satisfied, there may be regions where the performance is poor. We have demonstrated this with an example in Fig.~\ref{Fig:justification-for-K}. With $K=1$ (left), an overall success probability of $0.95$ is achieved in the entire work area using only $2$ demonstrations. However, resampling the work area with the same demonstrations but with a higher value of $K=16$ (right) shows that there exists a failure-prone region in the bottom right corner with a low individual success probability of $0.6$. The choice of $K$ is application specific, so that the requirement of being able to succeed (with high probability) in the overall work area alone can be met by choosing a smaller value of $K$. However, to identify the pockets of failure and to succeed in each individual region, one may need to choose a higher $K$, which in turn comes with the additional cost of a larger number of samples, as well as a few more demonstrations.

\section{Conclusion and Future Work}
This paper presents a novel approach to systematically obtain a sufficient set of kinesthetic demonstrations for complex manipulation tasks, one example at a time, such that the robot can develop a probabilistic confidence bound in its ability to generate feasible manipulation plans. Inspired by multi-arm bandit strategies, we propose an algorithm to partition the work area into disjoint subregions, ensuring successful plan generation in each region with a given probability that ultimately ensures overall success with high confidence.

The number of regions of interest for demonstrations (and the task space) increases exponentially with the number of task-relevant objects. Consequently, for the algorithm \texttt{getBestArm} (Alg.~\ref{alg:get-best-arm}) to scale well, we will need to use a combination of non-naive sampling methods such as adaptive sampling~\cite{auer2002finite}, and hierarchical methods to focus on smaller sets of promising candidates.

Finally, we are studying whether demonstrations given in one context can be reused in a new environment, i.e., a new work area or robot.  In this setting, it is possible to collect additional demonstrations in the new environment using the performance of the reused demonstrations as prior knowledge. Techniques for effectively reusing demonstrations in different environments remain an area to be further explored.

\ignore{Furthermore, we plan to explore another direction where we obtain a set of sufficient demonstrations in a certain region of the work area and use the associated probability distribution as a prior for Bayesian inference technique to obtain a new set of sufficient demonstrations in a different region of the work area or in a different work-area altogether.}



\bibliographystyle{IEEEtran} 
\bibliography{self_evaluation}

\appendix
\subsection{Architectural Overview of the User-Study}
\label{app:architecture}

Fig.~\ref{fig:architectural-overview} shows the architectural overview of the experimental setup for the user study.

\begin{figure*}[!t]
    \vspace{-1\baselineskip}
    \centering
    \scalebox{1.0}{
        \resizebox{\textwidth + \columnsep}{!}{
            \begin{tikzpicture}[outer/.style={draw=gray,ultra thin,rounded corners,densely dashed, fill=gray!5,inner xsep=0ex,inner ysep=0pt,yshift=0ex,fit=#1}]
                \begin{scope}[name prefix = top-]
                    \node (A) at (0,1.3) {\includegraphics[width=1.6cm]{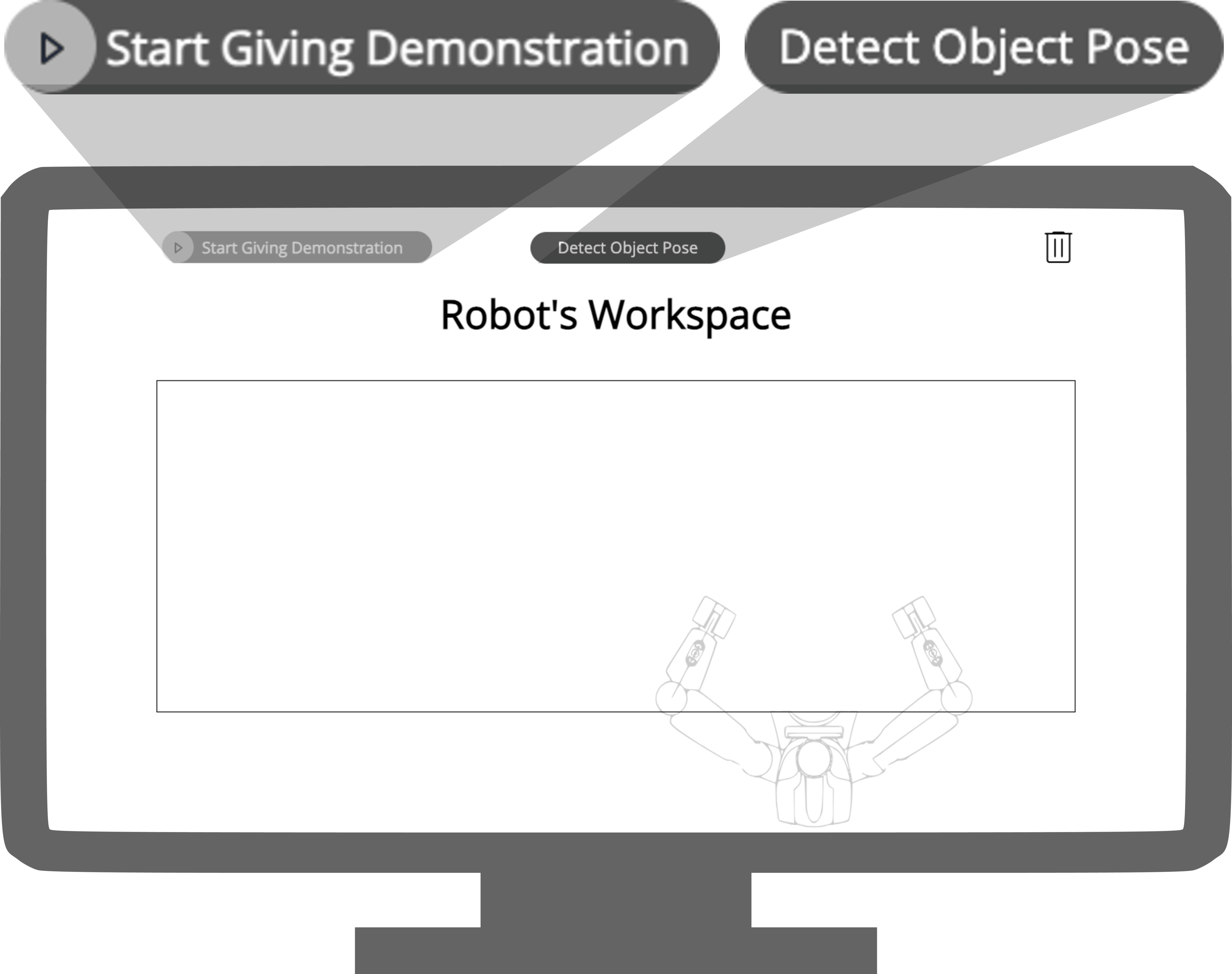}};

                    \node[yshift=11pt] at (A.south){\scalebox{0.25}{\textbf{[A]}}};
                    
                    \node[label={[yshift=6pt]below:\scalebox{0.26}{\begin{tabular}{@{}c@{}}
                       \textbf{[B]} Object Location for the\\ \emph{Initial} Demonstration
                     \end{tabular}}}] (B) at (1.7,1.3) {\includegraphics[width=1cm]{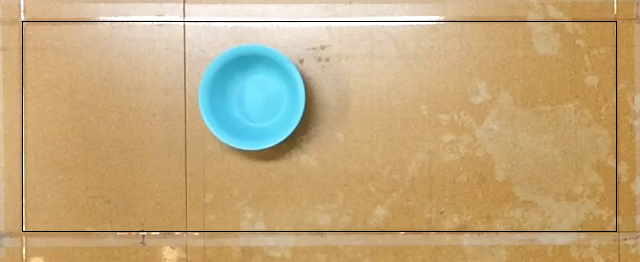}};
                
                    \node[label={[xshift=6.5pt,yshift=-6.5pt]above:\scalebox{0.25}{Object Detection}}] (BC1) at (2.2, 2.4) {\includegraphics[width=11mm]{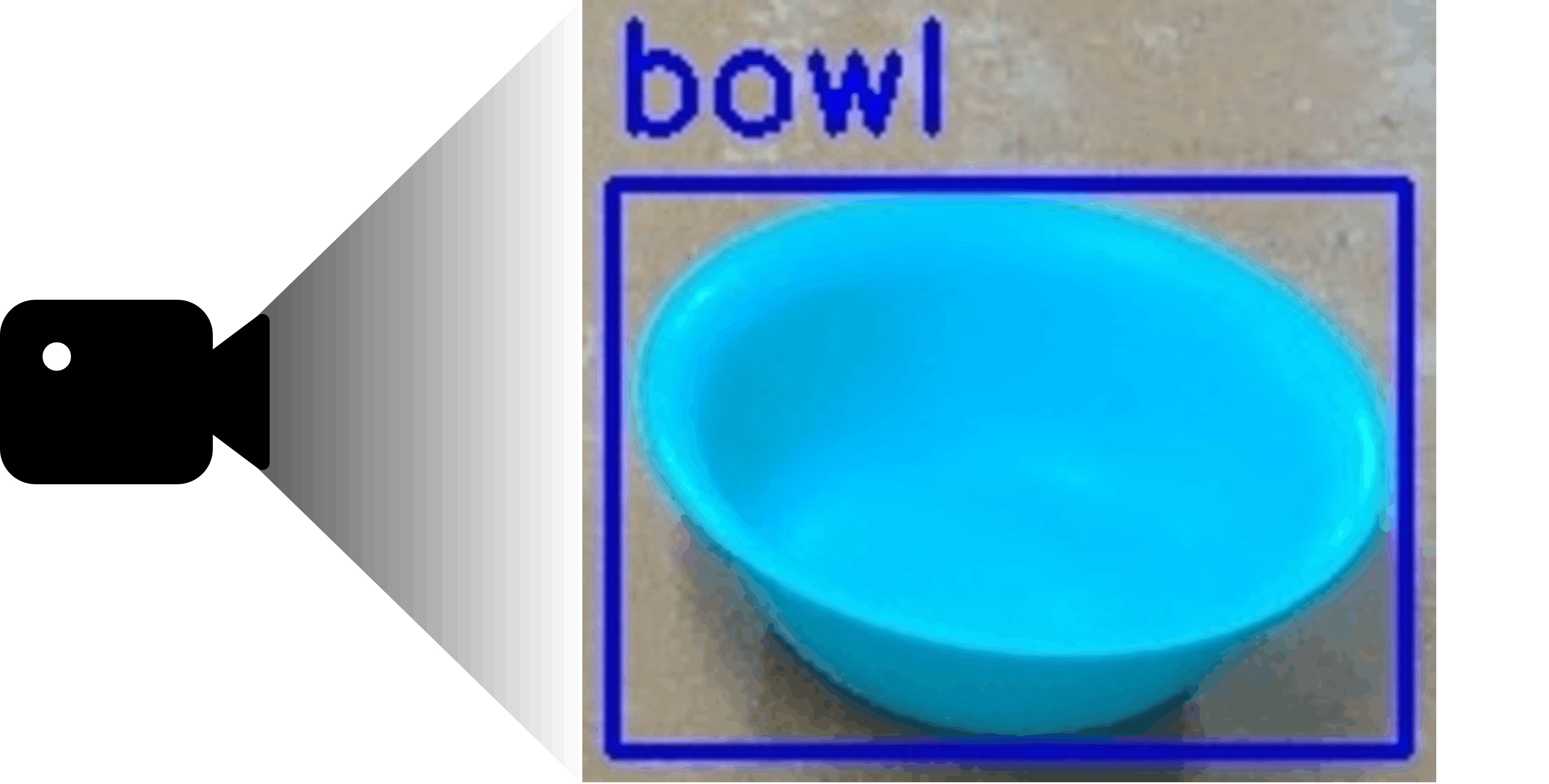}};
        
                    \node[label={[yshift=-5pt]above:\scalebox{0.25}{3D Point Cloud}}] (BC2) at (3.6, 2.4) {\includegraphics[width=6.5mm]{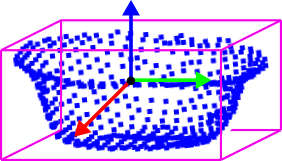}};
        
                    \scoped[on background layer]
                    \node[outer=(BC1) (BC2),label={[yshift=2pt]below:\scalebox{0.3}{Object Pose Estimation}}] {};
                
                    \node (C) at (4,1.3) {\includegraphics[width=1.6cm]{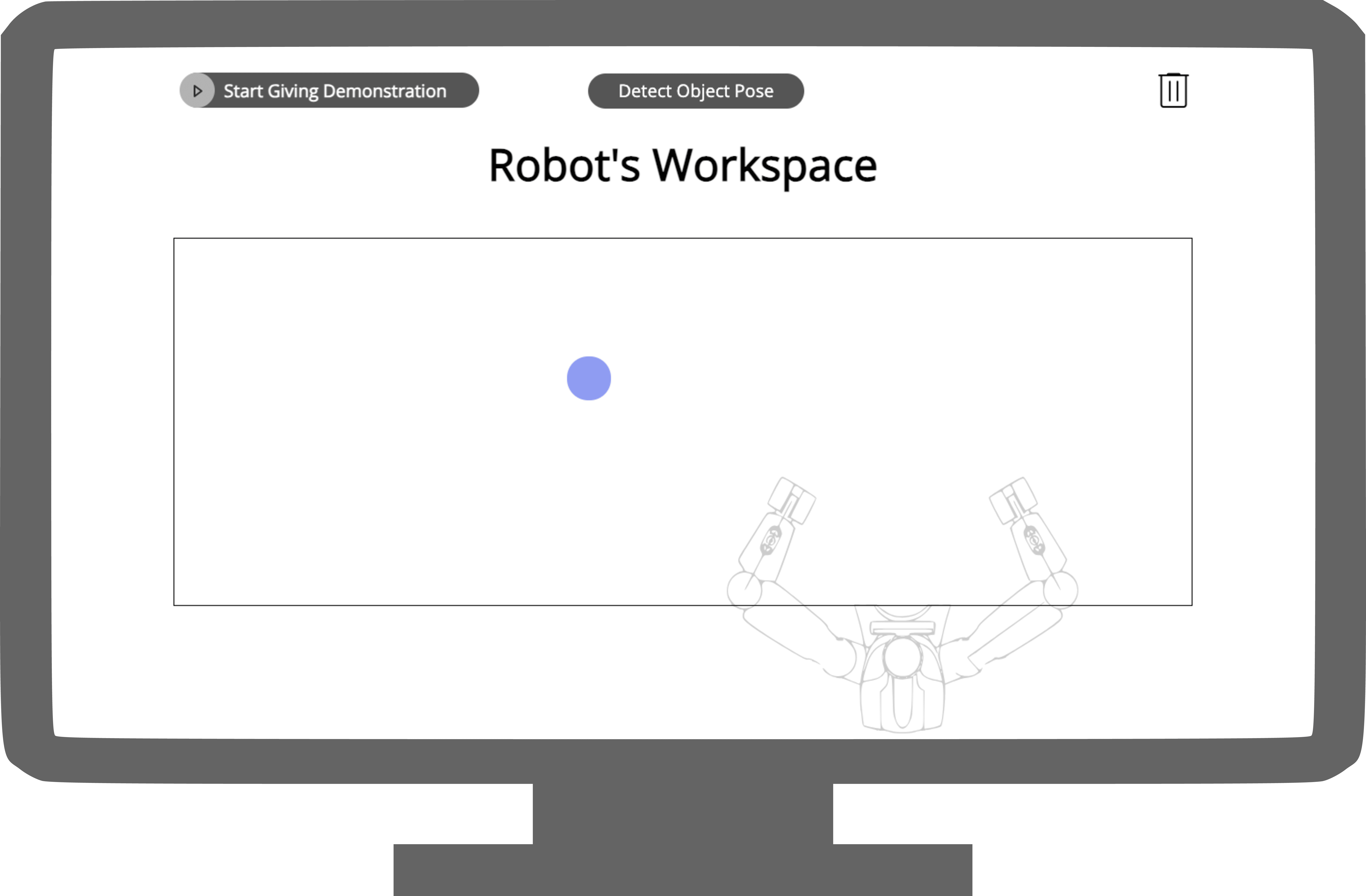}};

                    \node[yshift=11pt] at (C.south){\scalebox{0.25}{\textbf{[C]}}};

                    \draw[white,thick,line width=0.5pt,line cap=round,decoration={markings,mark=at position 1 with {\arrow{stealth'[darkgray!80]}}},postaction={decorate}] (A) to (B);

                    \draw[white,thick,line width=0.5pt,line cap=round,decoration={markings,mark=at position 0.57 with {\arrow{stealth'[darkgray!80]}}},postaction={decorate}] (B) to node[midway,above,yshift=0pt] {\includegraphics[width=9mm]{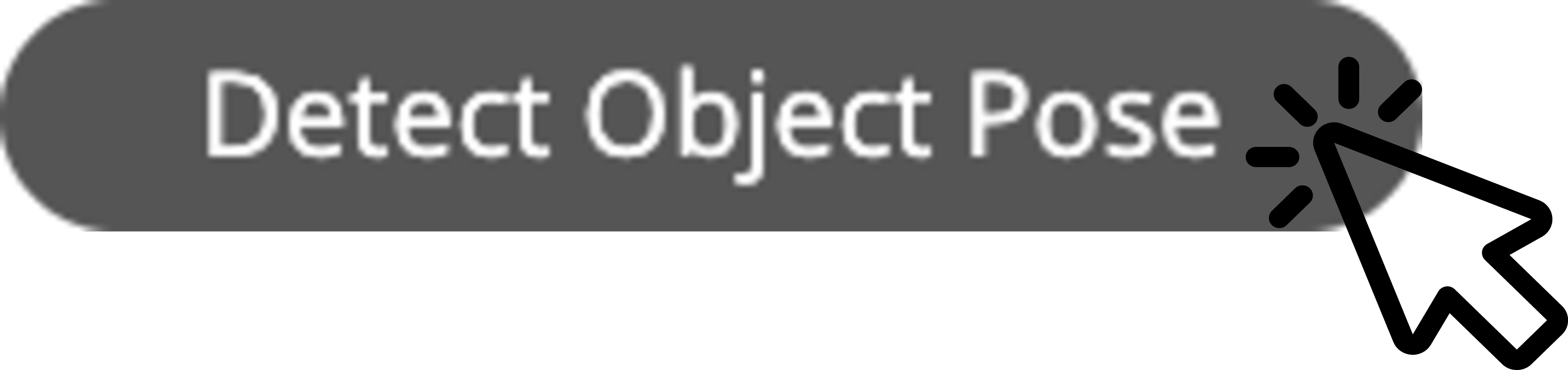}} (C);
        
                    \draw[blue!50,ultra thin,dotted,line width=0.5pt,line cap=round,decoration={markings,mark=at position 0.4 with {\arrow{stealth'[blue!50]}}},postaction={decorate}] ([xshift=5pt]B.north west) to [out=90,in=180] ([yshift=0pt]BC1.west);

                    \draw[gray!5,line width=0.5pt,line cap=round,decoration={markings,mark=at position 0.6 with {\arrow{stealth'[blue!50]}}},postaction={decorate}] (BC1) to (BC2);

                    \draw[blue!50,ultra thin,dotted,line width=0.5pt,line cap=round,decoration={markings,mark=at position 0.6 with {\arrow{stealth'[blue!50]}}},postaction={decorate}] ([yshift=6pt]BC2.east) to [out=0,in=90] ([xshift=-10pt]C.north east);
                \end{scope}
                
                \begin{scope}[name prefix = bottom-]
                    \node[label={[yshift=6pt]below:\scalebox{0.28}{\textbf{[D]} Demonstration Acquisition}}] (D) at (4,0) {\includegraphics[width=1.4cm]{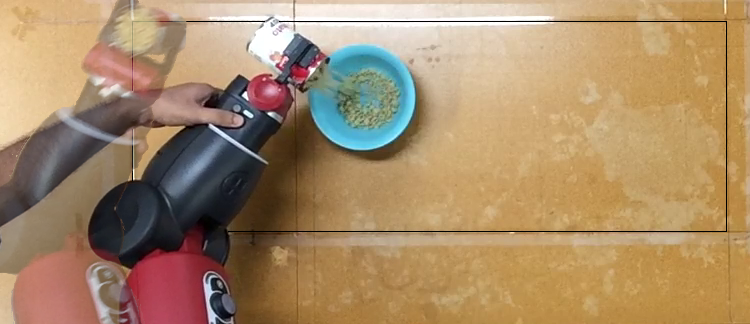}};
                    \node[xshift=-2pt,yshift=2pt] (E) at (1.6,0) {\includegraphics[width=1.6cm]{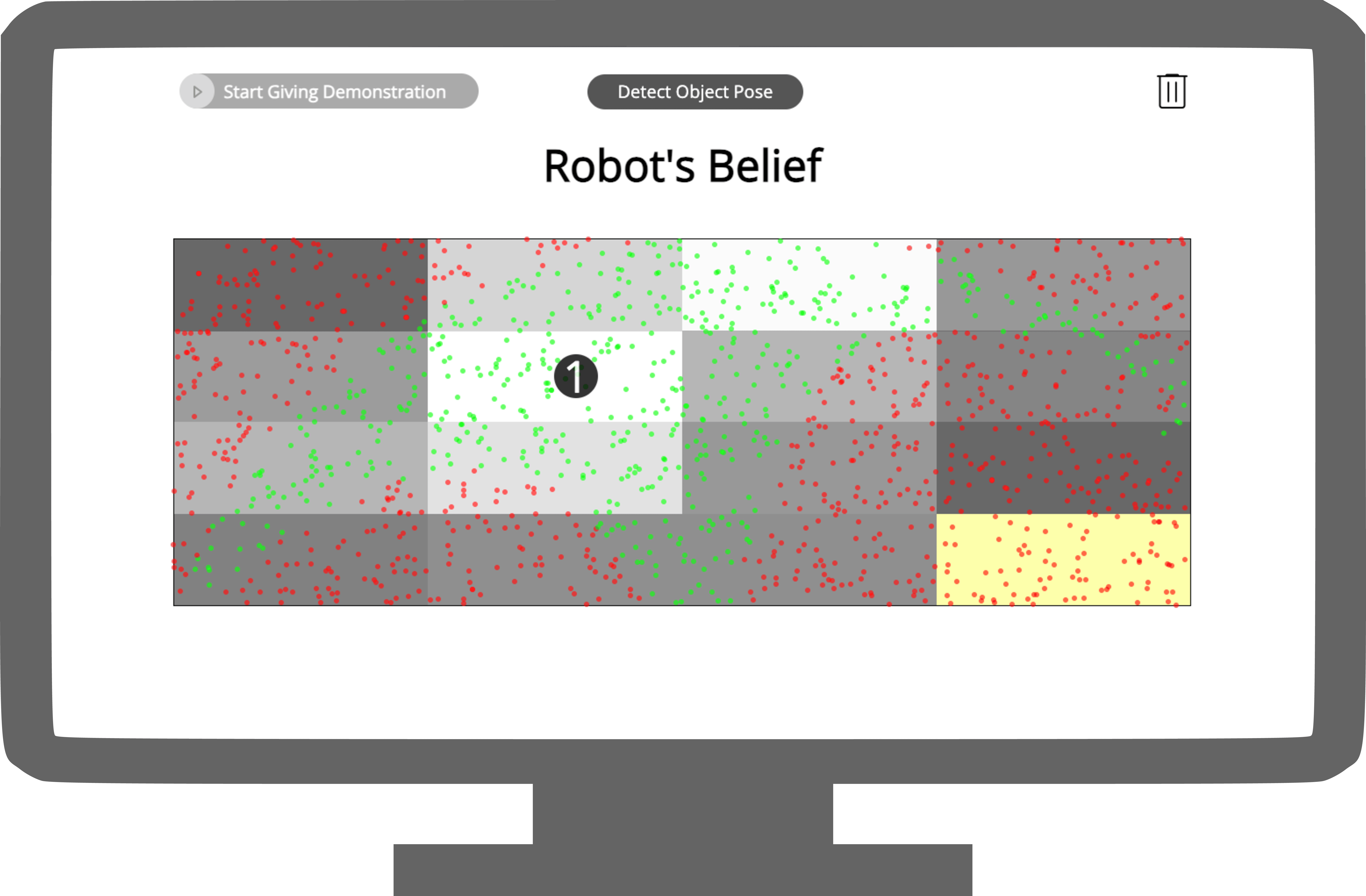}};

                    \node[yshift=11pt] at (E.south){\scalebox{0.25}{\textbf{[E]}}};

                    \node[label={[yshift=3pt]below:\scalebox{0.28}{
                     \begin{tabular}{@{}c@{}}
                       Motion segmentation and\\ ScLERP based Motion Plan Generation
                     \end{tabular}
                    }}] (DE1) at (3.6, -1.2) {\includegraphics[width=12mm]{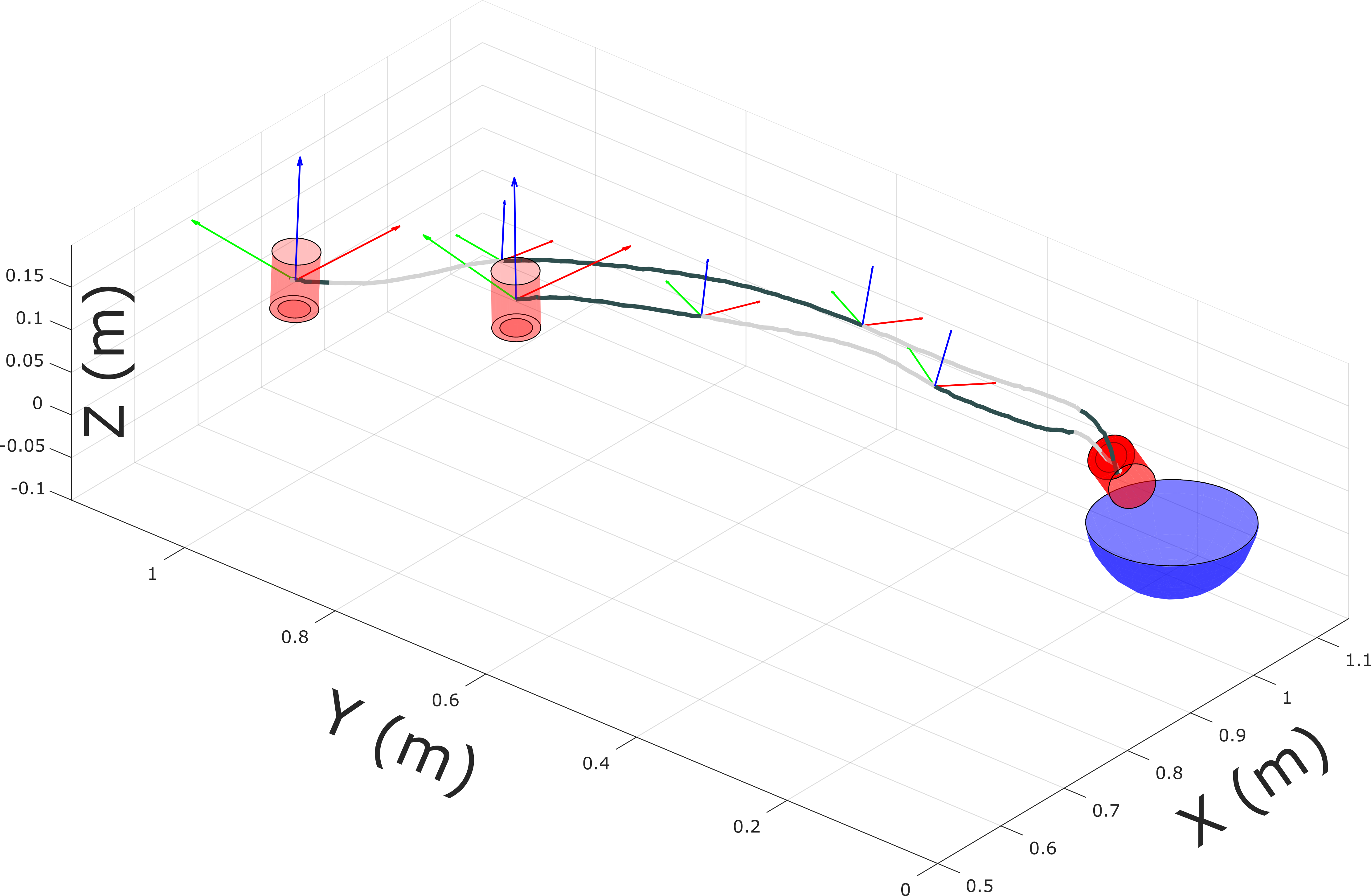}};
        
                    \node[label={[yshift=3pt]below:\scalebox{0.28}{
                     \begin{tabular}{@{}c@{}}
                       Multi-armed Bandit based\\ Belief Computation
                     \end{tabular}
                    }}] (DE2) at (1.8, -1.2) {\includegraphics[width=12mm]{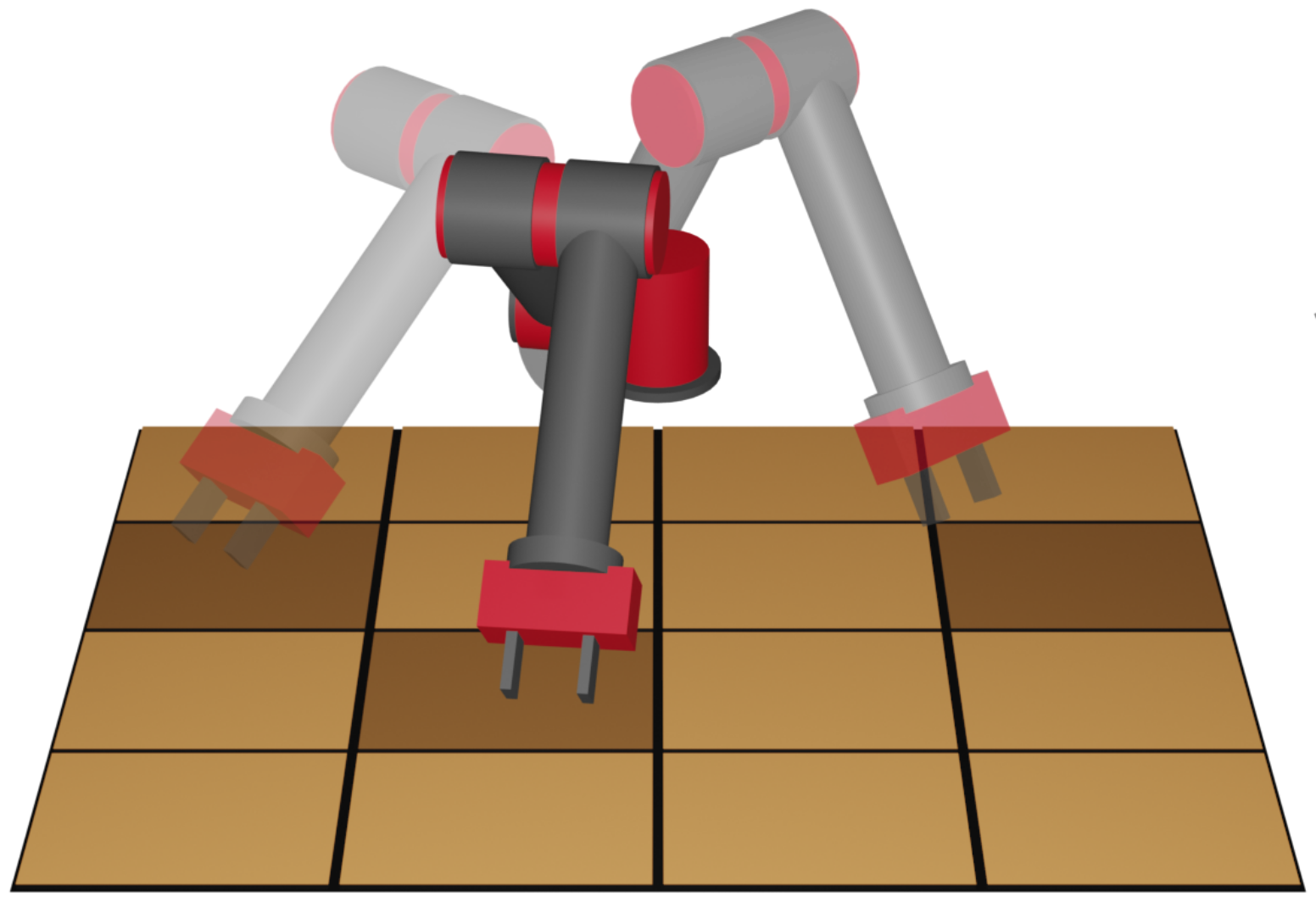}};
        
                    \scoped[on background layer]
                    \node[outer=(DE1) (DE2)] {};

                    \node[label={[yshift=6pt]below:\scalebox{0.26}{\begin{tabular}{@{}c@{}}
                       \textbf{[F]} Recommended Location\\ for the \emph{Next} Demonstration\\ (i.e. the location of the bowl)\\ within the Yellow Region
                     \end{tabular}}}] (F) at (-0.2,0) {\includegraphics[width=1cm]{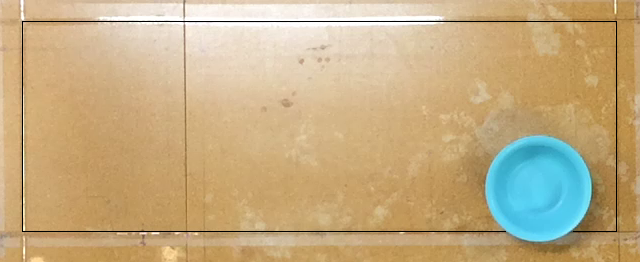}};

                    \draw[white,thick,line width=0.5pt,line cap=round,decoration={markings,mark=at position 0.6 with {\arrow{stealth'[darkgray!80]}}},postaction={decorate}] ([yshift=4pt]D.west) to node[midway,below,xshift=1pt,yshift=-6pt] (DE) {\includegraphics[width=10mm]{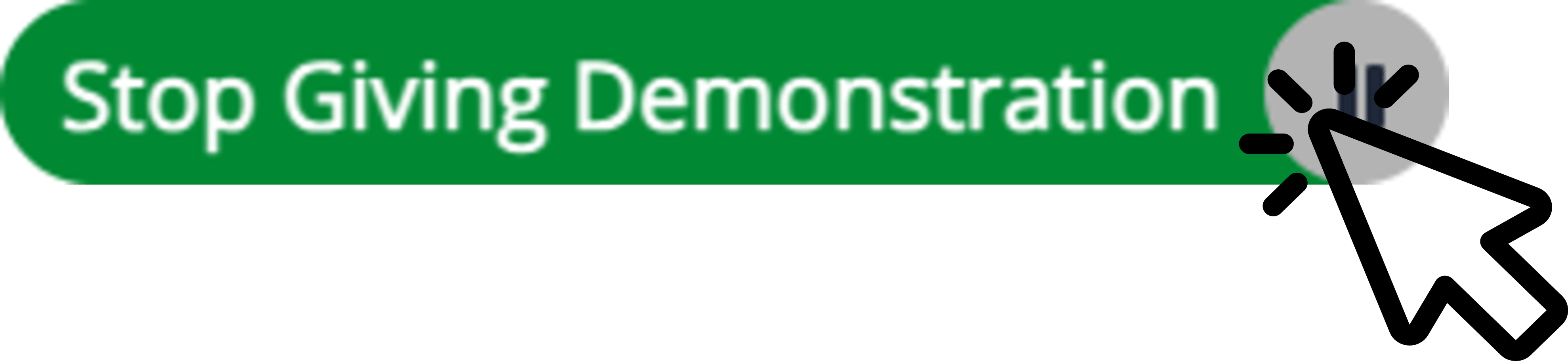}} (E);

                    \draw[white,thick,line width=0.5pt,line cap=round,decoration={markings,mark=at position 1 with {\arrow{stealth'[darkgray!80]}}},postaction={decorate}] (E) to (F);

                    \draw[blue!50,ultra thin,dotted,line width=0.5pt,line cap=round,decoration={markings,mark=at position 0.5 with {\arrow{stealth'[blue!50]}}},postaction={decorate}] ([xshift=-20pt,yshift=5pt]DE.south east) to [out=270,in=90] (DE2.north);

                    \draw[blue!0,ultra thin,dotted,line width=0.4pt,line cap=round,decoration={markings,mark=at position 0.7 with {\arrow{stealth'[blue!50]}}},postaction={decorate}] ([yshift=0pt]DE2.east) to [out=0,in=180] ([yshift=0pt]DE1.west);

                    \draw[blue!0,ultra thin,dotted,line width=0.4pt,line cap=round,decoration={markings,mark=at position 0.8 with {\arrow{stealth'[blue!50]}}},postaction={decorate}] ([yshift=-5pt]DE1.west) to [out=0,in=180] ([yshift=-5pt]DE2.east);

                    \draw[blue!50,ultra thin,dotted,line width=0.5pt,line cap=round,decoration={markings,mark=at position 0.6 with {\arrow{stealth'[blue!50]}}},postaction={decorate}] (DE2.west) to [out=180,in=270] ([xshift=8pt,yshift=6pt]E.south west);
                \end{scope}

                \draw[white,thick,line width=0.5pt,line cap=round,decoration={markings,mark=at position 0.65 with {\arrow{stealth'[darkgray!80]}}},postaction={decorate}] ([xshift=-4mm]top-C.south) to node[midway,right,xshift=1pt] {\includegraphics[width=10mm]{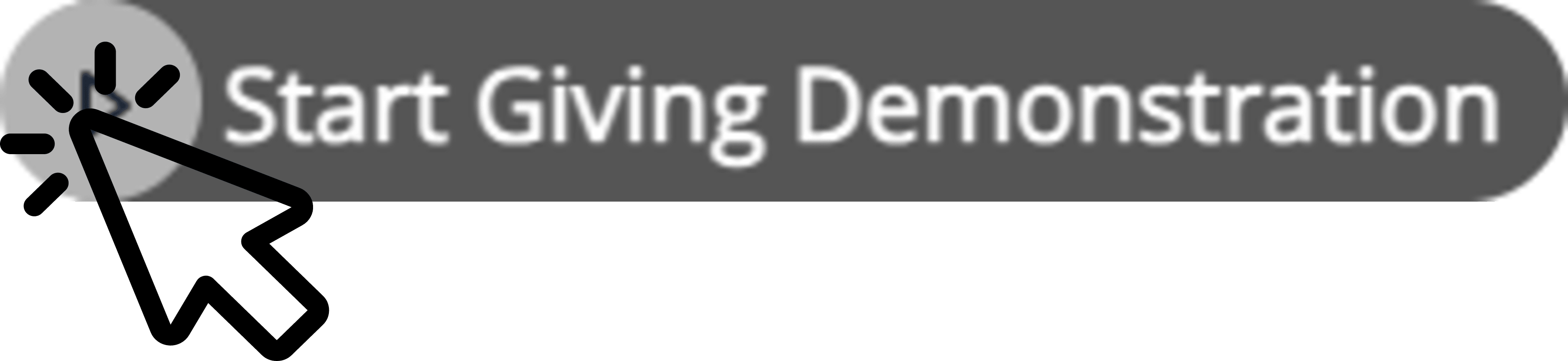}} ([xshift=-4mm]bottom-D.north);

                \draw[gray!25,dotted,thick,line width=0.5pt,line cap=round,decoration={markings,mark=at position 1 with {\arrow{stealth'[gray!25]}}},postaction={decorate}] (bottom-F) to node[black!80,midway,left,yshift=0pt] {\scalebox{0.28}{Repeat to \textbf{[C]} $\cdots$}} ([xshift=-5.5pt]top-A.south);
            \end{tikzpicture}
        }
    }

    \caption{Architectural workflow of DIRECT. The blocks in the middle (annotated from \textbf{[A]} to \textbf{[F]} with intermediate\ \ \protect\includegraphics[]{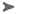} arrows) show the different stages of the interface (updated in real-time), as well as the corresponding real-world situations. The upper and the lower blocks (marked with\ \ \protect\includegraphics[]{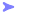} arrows) show the different computational modules used to compute the information rendered on the interface.}
    \label{fig:architectural-overview}
    \vspace{-0.5\baselineskip}
\end{figure*}

\subsection{Identification of an $\epsilon$-Optimal Arm in $K$-Arm Bandit}
\label{app:best-arm}

Given $K$ arms with unknown binary reward distributions $R_j$, $j \in [1,K]$, where $\mu_j = \mathbb{E}[R_j]$, the objective is to identify an arm `${a}$' which is $\epsilon$-close to the optimal arm, i.e. $\displaystyle\max_j \{\mu_j\} - \mu_a \leq \epsilon$.

Alg.~\ref{alg:naive_pac} gives a natural way to detect the arm with the highest expected reward by sampling each arm an equal number of times, say $\frac{T}{K}$, for a given number of samples $T$, and outputs the arm with the best empirical mean.

\begin{algorithm}
    \caption{$\epsilon$-optimal arm identification in K-arm bandit}\label{alg:naive_pac}
    \begin{algorithmic}
    \State Step 1: Collect $\left\lfloor\frac{T}{K}\right\rfloor$ samples from each of the $K$ arms
    \State Step 2: Calculate empirical means $\hat{\mu}_1, \hat{\mu}_2, \cdots, \hat{\mu}_K$ and output $\displaystyle\arg\max_j \{\hat{\mu}_j\}$
    \end{algorithmic}
\end{algorithm}

\begin{theorem}
Alg.~\ref{alg:naive_pac} is correct if $\lvert \mu_j - \hat{\mu}_j \rvert \leq \epsilon/2 \quad\forall j\in[1,K]$.
\end{theorem}
\begin{proof}
We assume `$\ast$' is the unidentified optimal arm, that is, $\forall j\ \ \mu_{\ast} \geq \mu_j$ and `$a$' is the arm found by Alg.~\ref{alg:naive_pac}, that is, $\forall j\ \ \hat{\mu}_a \geq \hat{\mu}_j$. We need to show that $\mu_{\ast} - \mu_a \leq \epsilon$.

\noindent Only $\hat{\mu}_a \in [\hat{\mu}_{\ast}, \hat{\mu}_\ast+\epsilon]$ (the hatched interval below) is consistent with our previous assumption as observed below.

\begin{tikzpicture}
  \draw[stealth'-stealth',color=gray] (-5,0) -- (3,0);
  
  \draw[shift={(-4,0)},color=red] (0pt,0pt) -- (0pt,3pt) node[above] {$\hat{\mu}_{\ast}-\frac{\epsilon}{2}$};
  \draw[shift={(-2,0)},color=red] (0pt,0pt) -- (0pt,3pt) node[above] {$\hat{\mu}_{\ast}$};
  \draw[shift={(0,0)},color=red] (0pt,0pt) -- (0pt,3pt) node[above] {$\hat{\mu}_{\ast}+\frac{\epsilon}{2}$};
  \draw[shift={(2,0)},color=red!40] (0pt,0pt) -- (0pt,3pt) node[above] {$\hat{\mu}_{\ast}+\epsilon$};

  \draw[shift={(-3.8,0)},color=blue] (0pt,0pt) -- (0pt,-3pt) node[below] {$\hat{\mu}_a-\frac{\epsilon}{2}$};
  \draw[shift={(-1.8,0)},color=blue] (0pt,0pt) -- (0pt,-3pt) node[below] {$\hat{\mu}_a$};
  \draw[shift={(0.2,0)},color=blue] (0pt,0pt) -- (0pt,-3pt) node[below] {$\hat{\mu}_a+\frac{\epsilon}{2}$};

  \path [pattern=north east lines, pattern color=gray!100] (-2,1.5pt) rectangle (2,-1.5pt);

\end{tikzpicture}

Thus, from the above diagram it is obvious that, $\hat{\mu}_a-\frac{\epsilon}{2} \leq \mu_a \leq \mu_{\ast} \leq \hat{\mu}_{\ast}+\frac{\epsilon}{2}$. Therefore, the difference between $\mu_{\ast}$ and $\mu_a$ is the maximum when $\mu_a = \hat{\mu}_a-\frac{\epsilon}{2}$ and $\mu_{\ast} = \hat{\mu}_{\ast}+\frac{\epsilon}{2}$ and the maximum difference is $\epsilon$.

Hence, $\mu_{\ast} - \mu_a \leq \epsilon$ that is, if $\lvert \mu_j - \hat{\mu}_j \rvert \leq \epsilon/2\quad\forall j\in[1,K]$, Alg.~\ref{alg:naive_pac} always returns an $\epsilon$-optimal arm.
\end{proof}

\begin{theorem}
Alg.~\ref{alg:naive_pac} returns an $\epsilon$-optimal arm with a probability of at least $1 - 2Ke^{\frac{-T\epsilon^2}{2K}}$
\label{hoeffding-theorem}
\end{theorem}

\begin{proof}
Let $E_j$ be the event $\lvert \mu_j - \hat{\mu}_j \rvert \leq \epsilon/2$. Thus, Alg.~\ref{alg:naive_pac} returns the correct arm, i.e. an $\epsilon$-optimal arm `$a$' with probability
\begin{gather}
    \textstyle\mathbb{P}\left(\bigcap_{i=1}^K E_j\right) = 1 - \mathbb{P}\left(\bigcup_{i=1}^K E_j^c\right) \left[E_j^c \text{ is complement of } E_j\right]  \nonumber\\
    \textstyle\ge 1 - \sum_{i=1}^K\mathbb{P}\left(E_j^c\right) \left[\because \mathbb{P}\left(\bigcup_{i=1}^K A_i \right) \le \sum_{i=1}^K \mathbb {P}(A_i)\right] \label{eq:1}
\end{gather}

Using the Hoeffding-Chernoff inequality for Bernoulli random variable $R_j$ with true mean $\mu_j$ and estimated mean $\hat{\mu}_j$ determined using $\frac{T}{K}$ samples, we get
\begin{align}
    \mathbb{P}\left(\lvert \mu_j - \hat{\mu}_j \rvert \le \frac{\epsilon}{2}\right) &\ge 1 - 2e^{\frac{-2T\epsilon^2}{4K}} \quad \forall j \in [1,K]\nonumber\\
    \mathbb{P}\left(E_j\right) &\ge 1 - 2e^{\frac{-T\epsilon^2}{2K}} \nonumber\\
    \mathbb{P}\left(E_j^c\right) &\le 2e^{\frac{-T\epsilon^2}{2K}} \label{eq:2}
\end{align}

Substituting~\eqref{eq:2} in~\eqref{eq:1}, we get $\mathbb{P}\left(\bigcap_{i=1}^K E_j\right) \ge 1 - 2Ke^{\frac{-T\epsilon^2}{2K}}$. Hence, Alg.~\ref{alg:naive_pac} returns an $\epsilon$-optimal arm with a probability of at least $1 - 2Ke^{\frac{-T\epsilon^2}{2K}}$.
\end{proof}

\begin{corollary}
For Alg.~\ref{alg:naive_pac} to succeed with a probability of at least $1 - \delta$, the number of samples $T$ should be at least $\frac{2K}{\epsilon^2}\ln\left(\frac{2K}{\delta}\right)$.
\end{corollary}

Substituting $2Ke^{\frac{-T\epsilon^2}{2K}}$ by $\delta$ in Theorem~\ref{hoeffding-theorem} gives us an ($\epsilon,\delta$)-PAC bound, in which, using at least $\frac{2K}{\epsilon^2}\ln\left(\frac{2K}{\delta}\right)$ samples we can identify an $\epsilon$-optimal arm with a confidence of at least $(1 - \delta)$.


\subsection{Screw-Geometry Based Motion Planning}
\label{app:sclerp-motion-planning}

\begin{figure}[!ht]
    \centering
    \includegraphics[width=\linewidth]{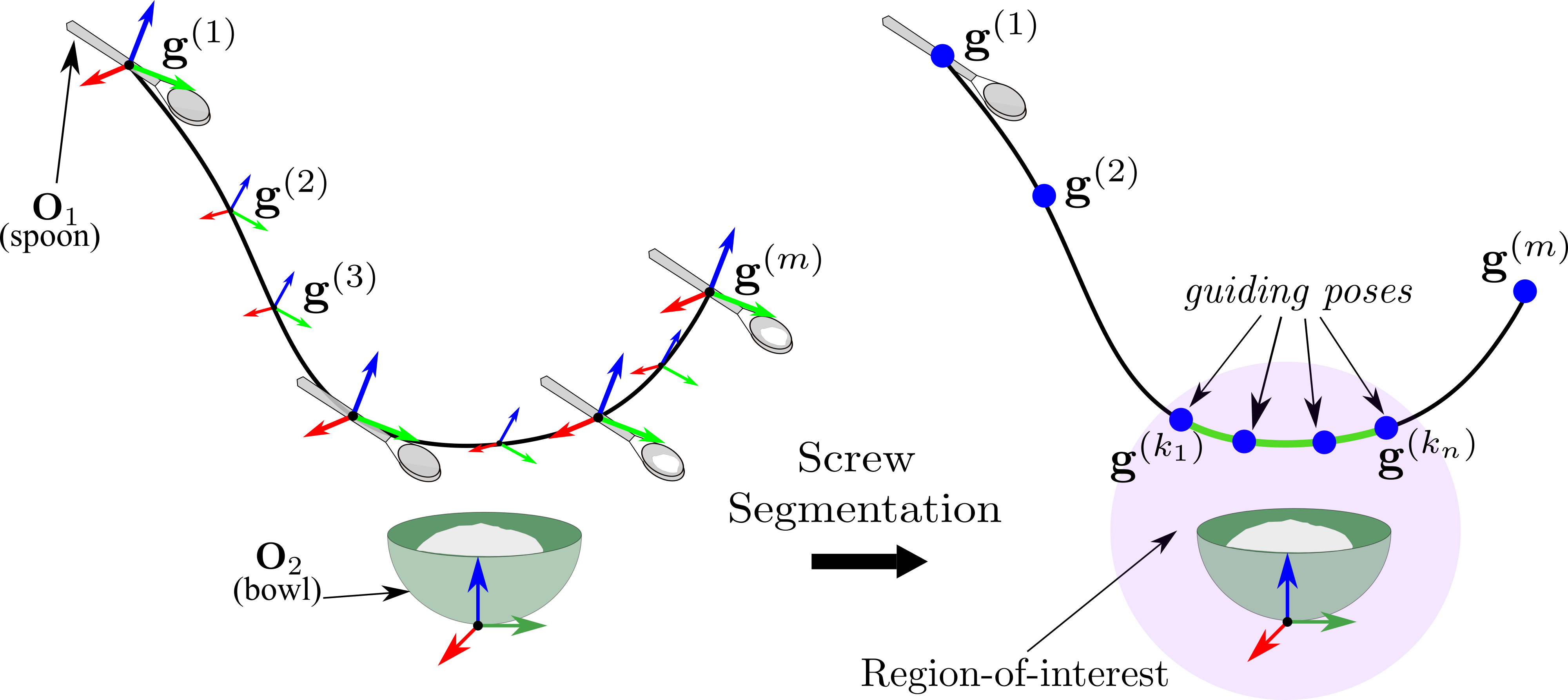}
    \caption{Overview of the motion segmentation algorithm \cite{mahalingam2023human} for scooping the contents from a bowl using a spoon. {\bf Left}: The human-provided demonstration is shown as a sequence of poses (of the spoon) in $SE(3)$, along with the pose of the bowl. {\bf Right}: Segmenting the sequence of poses in $SE(3)$ into a sequence of guiding poses. Each pair of consecutive guiding poses forms a constant screw segment. The task-relevant constraints are the sequence of constant screw segments in a region of interest surrounding the task-related object, bowl.}
    \label{Fig: Motion_Segmentation}
\end{figure}

For any manipulation task, there may be constraints on the end-effector's path that characterize the essential characteristics of the manipulation task. A kinesthetic demonstration of a task, given by holding the robot's arm in zero-gravity mode, traces an end-effector path that satisfies the task constraints. Let $\mathcal{J} \subset \mathbb{R}^d$ be the \emph{joint space}, i.e., the set of all possible joint configurations of the robot, where $d$ is the number of joints of the robot's manipulator arm. For a manipulation task instance, ${\bf y} \in \mathcal{X}$, a kinesthetic demonstration, $\bm{\Theta}$, is recorded as \emph{a discrete sequence of points from a joint space path}, i.e., a sequence of joint angle configurations $\bm{\Theta} = \left\langle \bm{\theta}^{(1)}, \cdots, \bm{\theta}^{(m)} \right\rangle$, where $\bm{\theta}^{(j)} \in \mathcal{J}$, $j = 1, 2, \cdots, m$, is a vector of dimension $d$, whose $i$-th component, $\theta_i^{(j)}$, represents the $i$-th joint angle of the robot arm. Every demonstration instance, $\bm{\Theta}$, is associated with a task instance ${\bf y}$. We keep this association implicit to avoid notational clutter.  \ignore{ Although there may be multiple demonstrations for a single task instance ${\bf y}$, for simplicity we can assume that there is only one $\bm{\Theta}_{\bf y}$ corresponding to each task instance ${\bf y}$, without loss of generality.}

Using a single demonstration \emph{to represent the task space motion constraints in the joint space is statistically ill-posed} since the (position inverse kinematics) mapping from the task space to the joint space is multi-valued (especially for redundant manipulators). Typically, any joint space-based technique uses multiple demonstrations, and it is usually difficult to \emph{a priori} specify the number of demonstrations required to learn the task space constraints. Therefore, we use a task space representation of the demonstration, which allows us to extract the motion constraints in the task space from a single demonstration by exploiting the screw-geometric structure of motion~\cite{mahalingam2023human}.

More specifically, following~\cite{mahalingam2023human}, each joint configuration, $\bm{\theta}^{(j)}$, is mapped to a pose of the robot's end effector frame, ${\bf g}^{(j)} \in SE(3)$, using position forward kinematics. Thus, the demonstration is a path, i.e., a sequence of poses in $SE(3)$, given by $\mathcal{G} = \left\langle {\bf g}^{(1)}, \cdots, {\bf g}^{(m)} \right\rangle$. By decomposing the path $\mathcal{G}$ into a sequence of constant screw segments, we obtain the representation of the demonstration as a sequence of guiding poses $\bm{\Gamma} = \left\langle {\bf g}^{(1)}, \cdots, {\bf g}^{(k_1)}, {\bf g}^{(k_2)}, \cdots, {\bf g}^{(m)} \right\rangle$. Figure~\ref{Fig: Motion_Segmentation} shows a schematic sketch of this process, and the details are presented in~\cite{mahalingam2023human}. Note that $\bm{\Gamma}$ is a subsequence of $\mathcal{G}$ and two consecutive poses in $\bm{\Gamma}$ represent a constant screw motion or a one-parameter subgroup of $SE(3)$.

The screw-geometric representation exploits the physical structure of motion implied by Chasles' theorem~\cite{murray2017mathematical} that {\em any path in $SE(3)$ can be arbitrarily closely approximated by a sequence of constant screw segments or one-parameter subgroups of $SE(3)$}. Thus, even one example is sufficient to extract the constant screw segments. \ignore{Furthermore, since $\bm{\Gamma}$ is a subsequence of $\mathcal{G}$, the constant screw representation naturally filters out some noise in the demonstration.}


\end{document}